\documentclass[10pt,twocolumn,letterpaper]{article}
\pdfoutput=1
\usepackage[pagenumbers]{cvpr} %
\usepackage{makecell}
\usepackage{tabularx}

\definecolor{cvprblue}{rgb}{0.21,0.49,0.74}
\usepackage[pagebackref,breaklinks,colorlinks,citecolor=cvprblue]{hyperref}
\usepackage{ragged2e}
\usepackage{multirow}
\usepackage{tabularx}
\usepackage{tabularray}

\title{3D Face Reconstruction From Radar Images}

\author{Valentin Braeutigam\textsuperscript{1}\\
\and
Vanessa Wirth\textsuperscript{1}\\
\and
Ingrid Ullmann\textsuperscript{1}\\
\and
Christian Schüßler\textsuperscript{1}\\
\and
Martin Vossiek\textsuperscript{1}\\
\and
Matthias Berking\textsuperscript{1}\\
\and
Bernhard Egger\textsuperscript{1}\\
\and 
\textsuperscript{1}Friedrich-Alexander-Universität Erlangen-Nürnberg
}

\begin{document}
\maketitle

\begin{abstract}
    The 3D reconstruction of faces gains wide attention in computer vision and is used in many fields of application, for example, animation, virtual reality, and even forensics. 
    This work is motivated by monitoring patients in sleep laboratories. 
    Due to their unique characteristics, sensors from the radar domain have advantages compared to optical sensors, namely penetration of electrically non-conductive materials and independence of light. 
    These advantages of radar signals unlock new applications and require adaptation of 3D reconstruction frameworks. 
    We propose a novel model-based method for  3D reconstruction from radar images. We generate a dataset of synthetic radar images with a physics-based but non-differentiable radar renderer. 
    This dataset is used to train a CNN-based encoder to estimate the parameters of a 3D morphable face model.
    Whilst the encoder alone already leads to strong reconstructions of synthetic data, we extend our reconstruction in an Analysis-by-Synthesis fashion to a model-based autoencoder. This is enabled by learning the rendering process in the decoder, which acts as an object-specific differentiable radar renderer.
    Subsequently, the combination of both network parts is trained to minimize both, the loss of the parameters and the loss of the resulting reconstructed radar image. 
    This leads to the additional benefit, that at test time the parameters can be further optimized by finetuning the autoencoder unsupervised on the image loss. 
    We evaluated our framework on generated synthetic face images as well as on real radar images with 3D ground truth of four individuals. 
\end{abstract}

\section{Introduction}
The reconstruction of humans using alternative capturing systems other than optical sensors has become an increasingly studied field of research over the past few years~\cite{chen_immfusion_2023, zhao_through-wall_2019, yue_bodycompass_2020, zhang_mmfer_2023}. 
Among them, a category gaining increasing interest is millimeter wave (mmWave) radar, which offers a significant advantage over other methods. 
Due to its wavelength, it is capable of penetrating certain obstacles, for example, fabric~\cite{ahmed_security} or even walls~\cite{zhao_through-wall_2019}, that impede the view of the object of interest. 
The technology is already in widespread use at airports for scanning passengers for illegal items prior to flight. In 2019 172 airports in the United States of America used these scanners~\cite{airports_mmWave}. 
However, it has the potential for application in several other fields, where radar signals can be utilized for computer vision tasks such as recognition and reconstruction. \\
An example in which it has the potential to be of significant benefit in the future is the monitoring of patients in clinics and sleep laboratories. 
Radar imaging offers the benefit compared to optical sensors in that it is not reliant on light and is able to monitor patients in their beds without removing any pillows or bed sheets. 
Consequently, this avoids the need for the patient to leave their bed for some examinations and allows their body to be monitored during night. 
This potentially fosters automatic sleep monitoring over manual supervision in the future. \\
Preliminary work has already been conducted, for example, radar-based techniques to track the pose of the human body during sleep~\cite{yue_bodycompass_2020}. 
In this paper, we concentrate on 3D reconstruction of human faces with their identity and expression. 
The face conveys a lot of information via facial expressions, which can be reconstructed, making it a key factor for assessing the state of patients during the night. However, due to the variety of facial shapes, it is challenging to reconstruct faces accurately. %
One significant challenge that arises from radar reconstruction of faces is the dependency of the viewing angle of the radar system and the face that is captured since the reflected signals depend on the surface normals~\cite{wirth2024maroonframeworkjointcharacterization}. Therefore, not all parts of the face are visible in the radar images which makes them ambiguous, as we show in the Supplementary Material. \\
This paper presents a learning-based approach for reconstructing 3D faces from radar images, utilizing a 3D morphable face model (3DMM). 
To this end, we generate a synthetic radar image dataset from faces constructed from the Basel Face Model (BFM) 2019~\cite{gerig_bfm_2017}. 
Given the data, we train two architectures: an encoder that is trained fully-supervised and an autoencoder that combines our pre-trained encoder with a learned differentiable renderer, thereby imposing an additional form of supervision to our network and enabling optimization at test time. \\
Our contributions are as follows:
\begin{itemize}
    \item a model-based 3D face reconstruction that for the first time only operates on images computed from radar signals
    \item an end-to-end training by approximating the physics-based radar renderer with a neural network which is differentiable and can generate synthetic radar images faster
    \item a publicly available dataset containing 10,000 synthetic radar images of faces generated with the BFM 2019~\cite{gerig_bfm_2017} and their corresponding parameters
\end{itemize}
\section{Related Work}

\textbf{Face Reconstruction From RGB Images.}
In the case of conventional RGB images, the 3D reconstruction of human faces has already been widely investigated, with a multitude of approaches proposed to address this challenge as summarized by Egger et al.~\cite{egger20203d}. %
These methods employ either a face model as prior~\cite{deng_accurate_2019, sanyal_learning_2019} or directly estimate the points of the face~\cite{sela_unrestricted_2017}. 
Examples of commonly utilized publicly available face models are the BFM~\cite{bfm09, gerig_bfm_2017} or, the FLAME face model~\cite{li_learning_2017}, which are created from 3D scanned heads of a large group of people. 
The approaches can be further divided into two categories: learning-based approaches ~\cite{deng_accurate_2019, sanyal_learning_2019, sela_unrestricted_2017, tran_regressing_2017}, and learning-free approaches ~\cite{amberg_optimal_2011}. 
Some of these methods are based on landmarks or keypoints~\cite{amberg_optimal_2011, tran_regressing_2017}, while others do not rely on landmarks~\cite{sela_unrestricted_2017}, but utilize an image-to-image approach, which maps the input image to a depth and a face correspondence image. \\
In regard to our work, the learning-based method proposed by Chang et al. \cite{chang_expnet_2018, chang_faceposenet_2017} is of particular significance, as a part of our architecture is based on their findings on the reconstruction of faces from RGB images. 
They combine two ResNet-101 models~\cite{he2016deep}, one trained to predict shape parameters, another one to predict facial expression parameters, and an AlexNet model~\cite{2012AlexNet} which estimates pose parameters. 
Subsequently, they apply the parameters to the BFM 2017~\cite{gerig_bfm_2017} to construct a 3D mesh of the corresponding face.\\
Another method that motivated our work is presented by Tewari et al. ~\cite{tewari_mofa_2017}. 
They employ a convolutional neural network (CNN) encoder to predict the parameters of their parametric face model as well as camera and lighting parameters. A differentiable renderer is then utilized to generate an image from these parameters which completes the model-based autoencoder. Since there is no differentiable radar renderer available, and building one is highly non-trivial (if possible), we implement a comparable differentiable autoencoder with a learned renderer. 
The core benefit of such a model-based autoencoder is that it can be trained unsupervised on the reconstruction task using a photometric loss.
The work of Li et al.~\cite{chunlu_li_robust_2021} utilizes the same autoencoder principle and focuses on reconstruction under occlusion. %
\\
\textbf{Human Reconstruction From Radar Signals.}
In contrast to the reconstruction of objects in an optical setup, methods for reconstruction with radar signals rarely operate on images, but commonly on raw signals~\cite{chen_immfusion_2023,zhao_through-wall_2018, zhao_through-wall_2019,yue_bodycompass_2020,zhang_mmfer_2023}. 
A subfield that is investigated is human body reconstruction, where the body is obscured by obstacles or impaired vision due to darkness or weather influences~\cite{chen_immfusion_2023,zhao_through-wall_2018, zhao_through-wall_2019}. \\
Chen et al.~\cite{chen_immfusion_2023} utilize radar signals in conjunction with RGB videos to reconstruct a full-body model of humans in visibility-impairing weather, for example, rain or fog, where optical sensors perform insufficiently. 
Other authors present methods that employ solid obstacles in their evaluation and utilize electromechanical signals within the WiFi frequency range to reconstruct the full body of humans hidden from view by walls or clothes~\cite{zhao_through-wall_2018, zhao_through-wall_2019}. 
In order to achieve this, they utilize the SMPL body model~\cite{loper_smpl_2015} as a prior geometry and then utilize a network based on the concept of Mask R-CNN~\cite{he2017mask} to predict its shape. 
Yue et al.~\cite{yue_bodycompass_2020} reconstruct the pose of a human in the dark during sleep via WiFi signals. 
The objective of these works is to reconstruct the entire human body or to determine its overall pose, whereas we focus on human faces. \\
Zhang et al.~\cite{zhang_mmfer_2023} classify facial expressions with radar signals through the detection of facial muscle movements using a mmWave radar sytem. 
With their approach they are capable of detecting facial expressions in a constrained setting with an accuracy of 80.57\%. 
Our work goes one step further and reconstructs the whole 3D face. \\
Xie et al.~\cite{xie_et_al_mm3dFace} are the first to perform 3D face reconstruction on radar signals. The authors use a learning approach based on the ConvNeXt-model~\cite{liu2022convnet} to predict facial landmarks. They utilize a FLAME~\cite{li_learning_2017} model to combine the landmarks into a continuous face. In our approach, we do not use landmarks, but predict the parameters of a face model to predict the whole face at once and in addition propose an autoencoder framework which enables optimization in an Analysis-by-Synthesis fashion which further improves the reconstruction. \\
While all of the aforementioned methods directly operate on radar signals without reconstructing an image from the signal, some radar-based approaches operate on images.
Bräunig et al.~\cite{j_braunig_ultra-efficient_2023} present an approach for 3D reconstruction that uses the theory of frequency shift keying to increase the reconstruction speed of hands compared to classical approaches. The authors utilize point clouds for reconstruction and show their results on human hands. 
Schüßler et al.~\cite{schuessler_sign_lang} present a ResNet-based approach to classify hand poses, which are selected from the American sign language system. 
\\
In our work we combine image-based approaches proven useful on RGB images and radar imaging techniques. 

\section{Methods}
In the following, we introduce the individual components of our work, which include our radar capturing setup, synthetic radar image dataset, our proposed models for model-based 3D reconstruction and the used training parameters.

\subsection{Radar Imaging}
\label{sec:radar_capturing}
\begin{figure}
    \centering
    \includegraphics[width=0.4\linewidth]{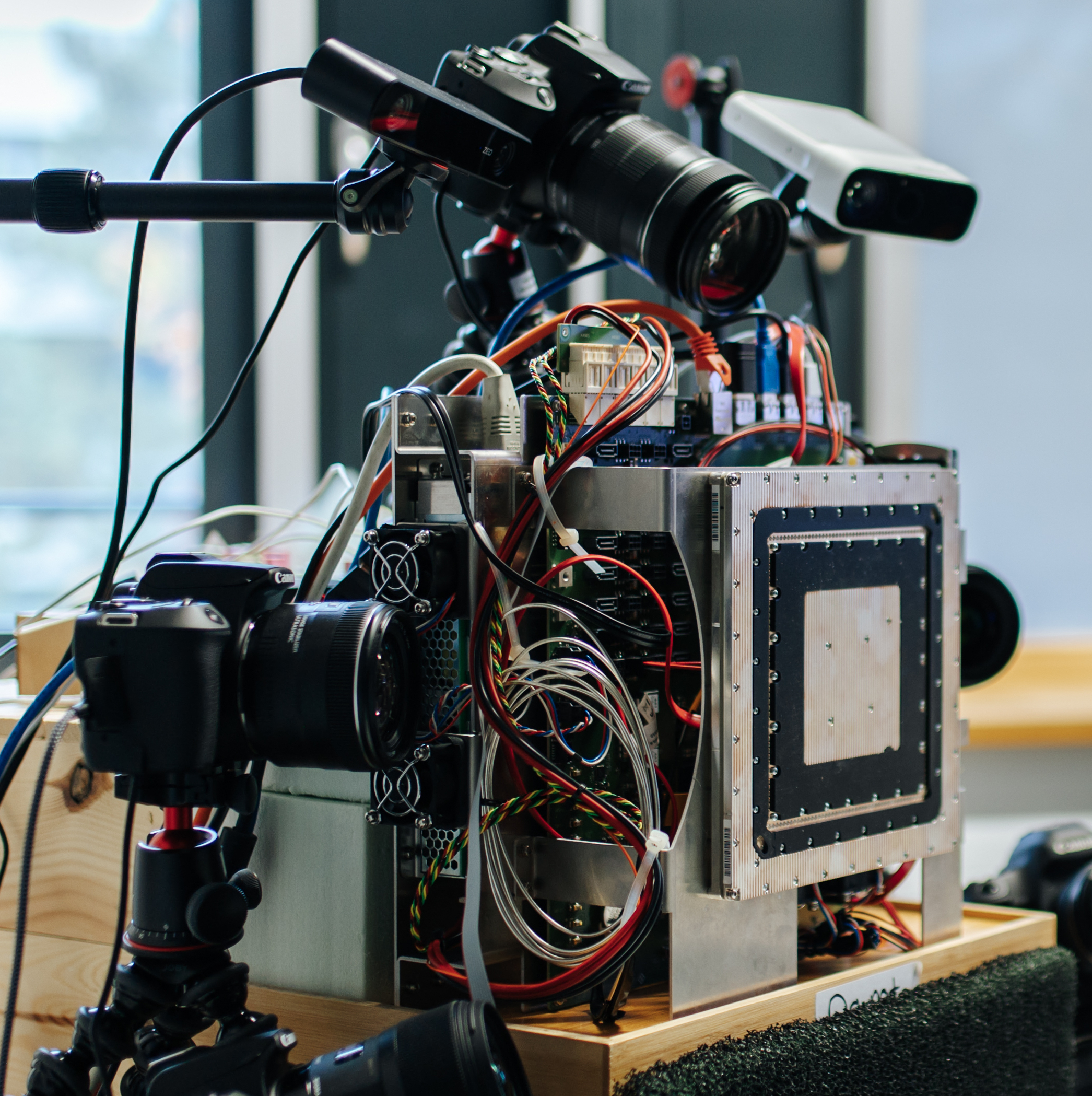}
    \includegraphics[width=0.535\linewidth]{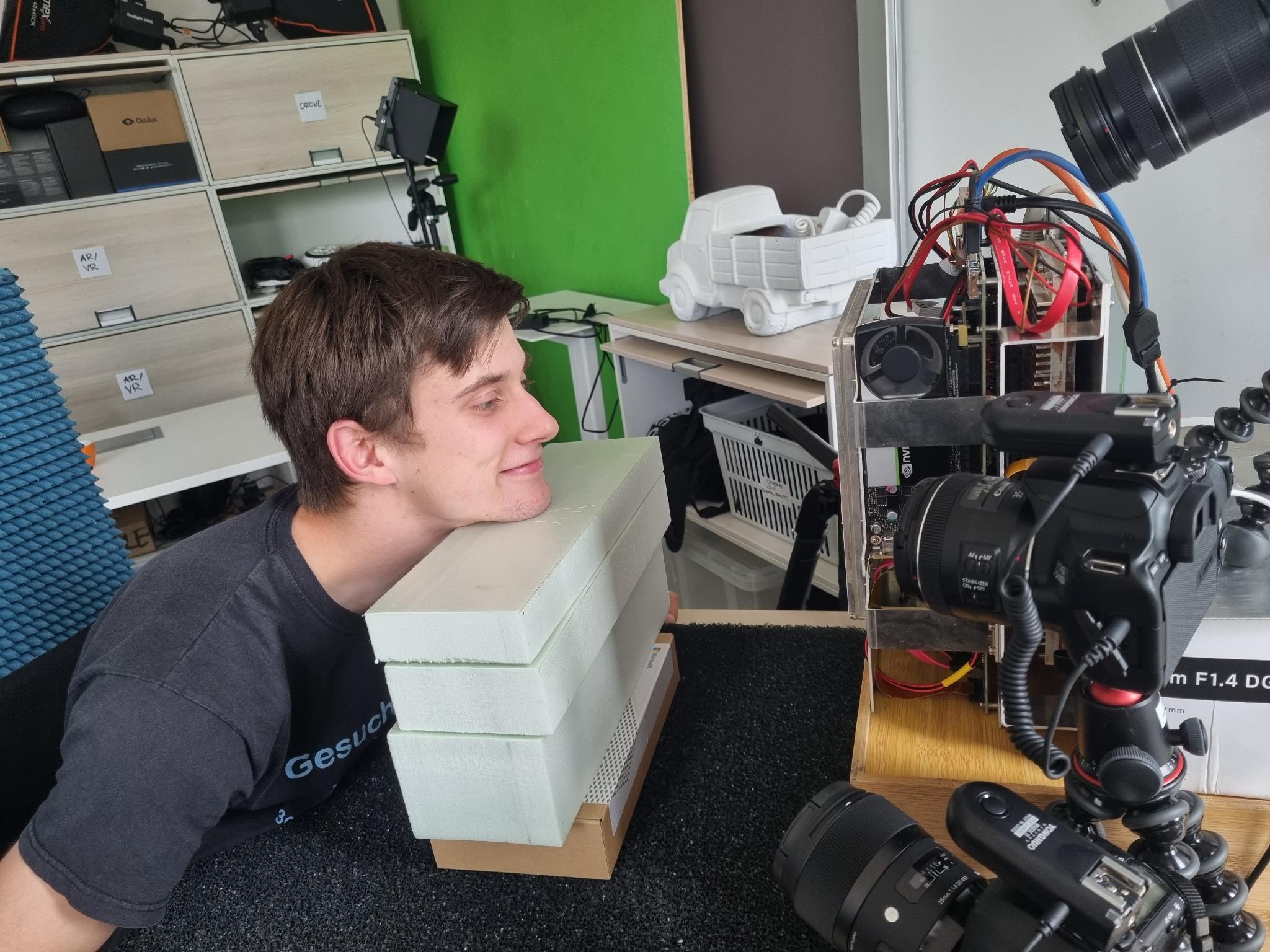} %
    \caption{The real radar setup and RGB cameras for photogrammetry. The radar module consists of 94 transmitter antennas and 94 receiver antennas in a square-shaped placement. Around the radar module five cameras are positioned to additionally reconstruct the captured face via photogrammetry. Four persons were captured in this setup, with each showing five different facial expressions. } %
    \label{fig:radar_setup}
\end{figure}
\begin{figure}
    \centering
    \begin{tabular}{p{0.05\linewidth} c}
       \begin{minipage}{\linewidth}
           \rotatebox{90}{Amplitude Image}
       \end{minipage} &  \begin{minipage}{0.95\linewidth}    \begin{subfigure}{\linewidth}
    \includegraphics[width=0.45\linewidth,trim={50px 30px 30px 50px},clip]{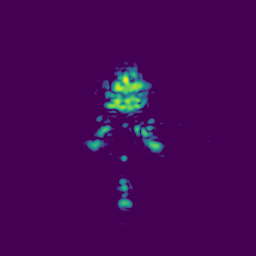}
    \hspace{-25pt}
    \includegraphics[width=0.06\linewidth]{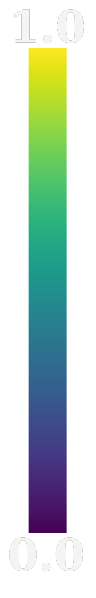}
    \hspace{5pt}
    \includegraphics[width=0.45\linewidth,trim={50px 30px 30px 50px},clip]{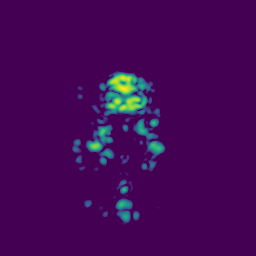}
    \hspace{-25pt}
    \includegraphics[width=0.06\linewidth]{images/workflow_figures/color_scheme_grey.pdf}
    \end{subfigure} \end{minipage}\\
      \begin{minipage}{\linewidth}
           \rotatebox{90}{Depth Image}
       \end{minipage}  & \begin{minipage}{0.95\linewidth} \begin{subfigure}{\linewidth}
    \includegraphics[width=0.45\linewidth,trim={50px 30px 30px 50px},clip]{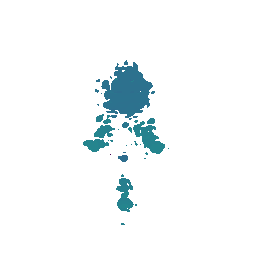}
    \hspace{-32pt}
    \includegraphics[width=0.12\linewidth]{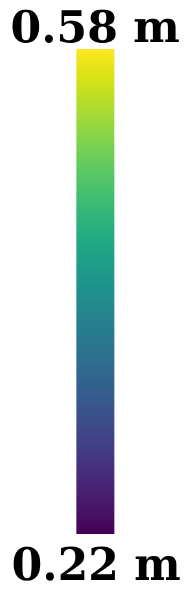}
    \hspace{5pt}
    \includegraphics[width=0.45\linewidth,trim={55px 30px 25px 50px},clip]{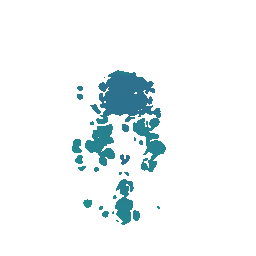}
    \hspace{-38pt}
    \includegraphics[width=0.12\linewidth]{images/workflow_figures/color_scheme_depth.pdf}
    \end{subfigure}
    \end{minipage}
    \end{tabular}

    \caption{Examples for a real radar images (left) and synthetic-real radar images (right). The amplitude images have a dynamic range of -20 dB. The synthetic real images are generated from the mesh of the same person reconstructed via photogrammetry.}
    \label{fig:comparison_real_synth}
\end{figure}

\textbf{Radar Setup.} We capture real radar images with the multiple-input multiple-output (MIMO) stepped-frequency continuous-wave (SFCW) radar system displayed in Figure~\ref{fig:radar_setup}. 
The system comprises 94 receiver antennas (RX) and 94 transmitter antennas (TX) in a square-shaped placement and is a submodule of an automotive radome tester~\cite{rohde_schwarz_QAR50, rohde_schwarz_imager}. 
The radar system uses frequencies from 72~GHz to 82~GHz in 128 equidistant steps. The spatial resolution of the radar system is approximately 4~mm in \textit{x}- and \textit{y}-direction and 11 mm in \textit{z}-direction. The face of the person being captured is located at a distance of 25~cm from the radar antennas. Behind the person are foam walls that minimize the reflection of radar signals behind the person. 
The aperture extent is approximately 14~cm~x~14~cm, with 3~mm spacing between the antennas. \\
\textbf{Radar Reconstruction.} The radar signals received by the radar system are reconstructed using a state-of-the-art approach for 3D mmWave image processing, namely back-projection~\cite{j_braunig_ultra-efficient_2023, ahmed_advanced_2012}. 
In the course of back-projection, for each RX and TX antenna combination, the correlation between the received signal and a signal hypothesis is summed in a 3D voxel grid~\cite{ahmed_security}. 
Subsequently, a 2D array is extracted from this 3D voxel grid via maximum projection~\cite{keller_angiography} along the z-axis and the values are scaled logarithmically since they range over several orders of magnitude. 
Henceafter, the dynamic range $\theta$ is restricted to remove noise induced by the radar signals. In our experiments we use a dynamic range of -15 dB for the decoder, but we sample from a range of values for encoder training. Afterwards, the range of the resulting data is scaled linearly to the range [0,1] to convert it to a radar image, which we call an amplitude image.  \\
\textbf{Multimodal Capture.} During the capturing of radar signals optical cameras are simultaneously employed to reconstruct the face via photogrammetry. 
Once the 3D reconstruction of the faces has been completed, they are used in the radar simulation, resulting in a synthetic version of the real face. 
The resulting images can be utilized for a quality comparison of the radar simulation. 
The capture setup configuration is illustrated in Figure~\ref{fig:radar_setup}. 
A comparison between the real amplitude image and the synthetically generated amplitude image from the photogrammetry mesh is shown in Figure~\ref{fig:comparison_real_synth}. \\
In order to evaluate our approach on real data, we reconstructed real radar images of four male european individuals with fair skin and generated synthetic radar images from the corresponding photogrammetry mesh. \\
\textbf{Depth Images.} We employ depth images generated based on the radar signal as an additional input. 
The depth of each pixel is computed from the brightest scatterer along each depth slice of the voxel grid. 
In our experiments, these depth images are utilized as an alternative to the amplitude images, and also in conjunction with the amplitude images as input to the model. 
\begin{figure}
    \centering
    \begin{subfigure}{\linewidth}
    \includegraphics[width=0.45\linewidth,trim={40px 40px 40px 40px},clip]{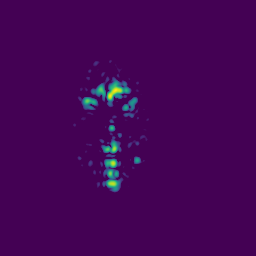}
    \hspace{-25pt}
    \includegraphics[width=0.06\linewidth]{images/workflow_figures/color_scheme_grey.pdf}
    \hspace{5pt}
    \includegraphics[width=0.45\linewidth,trim={40px 40px 40px 40px},clip]{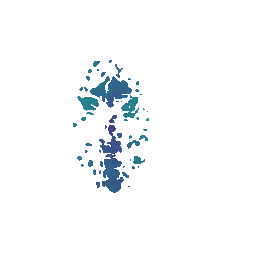}
    \hspace{-32pt}
    \includegraphics[width=0.12\linewidth]{images/workflow_figures/color_scheme_depth.pdf}
    \end{subfigure}
    \caption{Examples of a synthetic amplitude image (left) with a dynamic range of -20 dB and a synthetic depth image in comparison (right).}
    \label{fig:syn_images}
\end{figure}
\begin{figure*}
    \centering
    \includegraphics[width=0.95\linewidth]{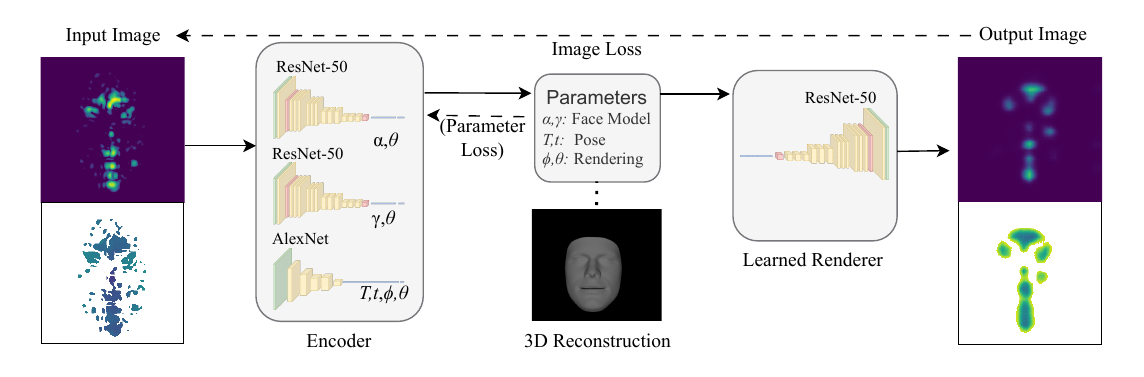}
    \caption{Overview of our method. The input image is fed to three encoder networks which predict the shape, expression, and pose of the face. 
    These parameters are then fed to a differentiable renderer that reconstructs the input image. 
    The encoder consists of two ResNet-50 models for predicting shape and expression and an AlexNet model for predicting the pose. 
    The differentiable renderer is a ResNet-50 model that is ordered in the opposite direction. During training, both the parameter loss and image loss are applied. For inference, the encoder and decoder are frozen and only the image loss is optimized leading to the face model parameters holding the 3D face reconstruction.}
    \label{fig:overview_image}
\end{figure*}

\subsection{Training Dataset}
\label{sec:dataset}
We generated a synthetic dataset of 10,000 instances of face meshes with varying facial expressions. 
To this end, the face mask \textit{face12} of the BFM 2019~\cite{gerig_bfm_2017} was utilized. \\
The generation of each face is based on a Gaussian-sampled shape vector $\alpha$ and expression vector $\gamma$ with a distribution of $ \mathcal{N}(0,1) $. 
The aforementioned vectors define the appearance of each face instance and the associated facial expression, as specified by the shape model. 
Additionally, the face is transformed by a uniformly sampled pose. 
The pose includes a translation $t$ in each \textit{x}- and \textit{y}-direction between \text{-5~cm} and \text{+5~cm}, the \textit{z}-value which is set to zero, rotations $T$ around the yaw axis between \text{-5} and 5 degrees, and the pitch and roll axes between \text{-10} and 10 degrees, respectively. 
For training and evaluation, the pose parameters are linearly scaled to values within [-1,~1]. \\
Subsequently, the aforementioned meshes are employed to generate synthetic radar images utilizing the radar renderer by Schüssler et al.~\cite{schuessler_et_al}, which was adjusted to generate images of human hands by Bräunig et al.~\cite{braeunig_et_al}. 
The renderer approximates the radar signals of a near-field MIMO radar system by raytracing and performs the back-projection algorithm~\cite{j_braunig_ultra-efficient_2023, ahmed_advanced_2012}, analogously to real radar signals. 
To model the reflection property Schüssler et al. use a material factor to interpolate linearly between a diffuse and a specular material. 
Another attribute is the size of the simulated antennas, which differs from the real antenna size to assimilate the difference between real radar waves and simulated rays. 
For our dataset, we sample these two rendering parameters $\phi$, the material factor within [0, 1], and the antenna size within [0.2, 0.3].  %
We then train the neural networks on these training images using the procedures outlined in the subsequent Sections~\ref{sec:encoder} and~\ref{sec:learned_renderer}. 

\subsection{Encoder}
\label{sec:encoder}
Two CNNs with the ResNet-50 \cite{he2016deep} architecture are employed for the prediction of shape and expression parameters of the face model. 
Furthermore, an AlexNet model \cite{2012AlexNet} is utilized to predict the pose. 
The outputs of the ResNet-50 models are scaled by applying a tanh-layer and multiplying the results with a scaling factor of three since this is the value range that contains 99.8 \% of the values sampled from the dataset while still reducing the possible value range. 
For all networks, we utilize an additional fully connected layer that outputs the expected amount of parameters. 
The encoder architecture is based on Chang et al.~\cite{chang_expnet_2018, chang_faceposenet_2017}, while we use smaller networks and do minor adjustments like the tanh-layer.\\
The aforementioned models are trained on a training set of 8,500 synthetic radar images and evaluated on the remaining 1,500 images. 
During training, we apply randomly sampled dynamic ranges between -15~dB and -30~dB, for evaluation we apply a fixed dynamic range of -20 dB. The models predict the first 3DMM parameters $\alpha \in \mathbb{R}^{10}$ and $\gamma \in \mathbb{R}^{7}$, which cover approximately 85\% of the shape variance and 76\% of the expression variance of the BFM, and the pose parameters $t \in \mathbb{R}^{3}$ and $T \in \mathbb{R}^{3}$ applied to the face model mesh. %
Additionally, the networks estimate the applied dynamic range, the material, and antenna size properties. 
During training phase, the L2 loss between those parameters and the resulting parameters is employed as a loss function.

\subsection{Learned Renderer and Autoencoder}
\label{sec:learned_renderer}
\textbf{Learned Renderer.} As we like to achieve an unsupervised optimization, we need a renderer that generates a radar image based on our parameters. Since the renderer we used for creating our dataset is not fast enough for the training and also not differentiable, we approximate the physics-based renderer through training a decoder network on generated images. Therefore, we use a ResNet-50, with its layers ordered in the opposite direction. A fully connected layer is employed to map the input to the first convolutional layer of the ResNet-50. This renderer is trained on the parameters to generate amplitude images with an applied dynamic range of -15 dB, and depth images, respectively. It utilizes the face model, pose and rendering parameters as described in Section~\ref{sec:dataset} to render synthetic radar images. We can then employ it to improve the training of our model. \\
\textbf{Autoencoder.}\label{sec:autoencoder} The learned renderer is combined with the encoder as displayed in Figure~\ref{fig:overview_image}. 
In the case of the autoencoder training, the pre-trained models of the encoder and the decoder are utilized, with the weights of the latter being fixed. This ensures that inductive constraints of the parameters in the latent space do not experience structural changes induced by the autoencoder, but rather remain aligned with the original intention of face model parameters. 
Subsequently, the autoencoder is trained on the synthetic image set. The prediction of face model parameters is combined with the task of reconstructing the input image from the parameters. \newline
The training loss is computed as follows:
\begin{equation}
     \mathrm{L}_{train} = \mathrm{L}_{image} + \lambda \cdot \mathrm{L}_{params},
\end{equation}
where $\mathrm{L}_{image}$ denotes the L2 loss between output and input image, $\mathrm{L}_{params}$ denotes the L2 loss between the params computed by the encoder part and the ground truth parameters. $\lambda$ is a weighting factor to adjust the importance of the loss functions relative to each other. Since the results have not changed substantially for other values of $\lambda$, we use $\lambda = 1$.
The network is trained on the same parameters as the encoder, described in the previous Section~\ref{sec:encoder}. \\
During evaluation both the encoder and the decoder are fixed and the latent space variables are further optimized by the image loss. 
\subsection{Model Training}
The Adam optimizer~\cite{kingma2017adammethodstochasticoptimization} is employed in the training of our network models, utilizing a Nvidia A100 graphics card. 
The encoder is trained with linearly scheduled learning rates in [0.01, 0.001], over the inital 150 epochs of the training period and over 200 epochs in total. 
To train the decoder and the autoencoder a learning rate of 0.001 is employed without learning rate scheduling over 300 epochs. 
We train the encoder and autoencoder in batches of 50 images, and the decoder in batches of 150 images. 
The autoencoder takes approximately 4 hours of training time, the encoder 2 hours and the decoder 1.5 hours, the training with radar and depth images combined takes 0.75 to 1 hour longer due to the higher amount of data. \\
\textbf{Runtime.} It takes 58~ms in the mean to get the resulting image from the trained decoder models, while the physics-based renderer used for the creation of the dataset takes about 2~min to reconstruct an image. Both measurements are done on a Nvidia GeForce RTX 4060 Ti. 

\section{Experiments \& Results}
\label{sec:results}
\begin{table*}
  \centering
  \begin{tabular}{c@{\hskip 12pt}c@{\hskip 12pt}c@{\hskip 12pt}c@{\hskip 12pt}c@{\hskip 12pt}c@{\hskip 12pt}c@{}}
    \toprule
    Method & L2 Shape $\downarrow$ & L2 Expr $\downarrow$ &  L2 Translation $\downarrow$ & L2 Rotation $\downarrow$ & L2 Total $\downarrow$ & Point Dist $\downarrow$\\
    \midrule
    Baseline (Mean) &  0.99053 &  0.99961 & 0.33157 & 0.22639 & 0.80767 & 4.42 mm\\
    Baseline (Random) &  2.01233 &  2.05933 & 0.67529 & 0.44667 & 1.64802 & 6.28 mm\\
    \midrule
    Encoder (Amplitude) & 0.83999 &  0.92118 & 0.06874	 & 0.00630 & 0.65536 & 3.47 mm \\
    Encoder (Depth) & 0.73976 & 0.91026 & 0.06458 & 0.00527 & 0.60778 & 2.77 mm\\
    Encoder (Amp.-Depth) & 0.73502 &  	0.88483 & \textbf{0.06128} & \textbf{0.00510} & 0.59753 & 2.82 mm\\
    \midrule
    Autoencoder (Amplitude) & 0.77045 &  \textbf{0.81509} & 0.07939 & 0.00704 & 0.59432 & 3.29 mm \\
    Autoencoder (Depth) & \textbf{0.63383} & 0.84990 & 	0.06570 & 0.00617 & 0.54362 & \textbf{2.56 mm} \\
    Autoencoder (Amp.-Depth) & 	0.63999 & 0.82016 & 0.07792 & 0.00713 & \textbf{0.53897} & 2.61 mm \\
    \bottomrule
  \end{tabular}
  \caption{L2 error of the parameters computed by the encoder and the autoencoder evaluated on the synthetic validation set and the mean point distance of all corresponding mesh points. The mean L2 error for the evaluated parameters is given and split into the shape and expression parameters by the face model and the translation and rotation of the face mesh. The point distance is computed by the mean Euclidean distance of each point pair of the resulting mesh and the ground truth mesh, without applying the pose. The models are trained on synthetic amplitude images, depth images, and amplitude-depth images. The estimation of the pose has similar quality for all approaches, while the autoencoder performs better for shape and expression estimation and has a lower point distance, which are the core tasks of interest. The models trained on depth or amplitude-depth input perform better in the shape estimation and have a lower point distance.}
  \label{tab:loss_comparison}
\end{table*}

\begin{figure}
    \begin{subfigure}[t,c]{\linewidth}
    \includegraphics[width=0.82\linewidth]{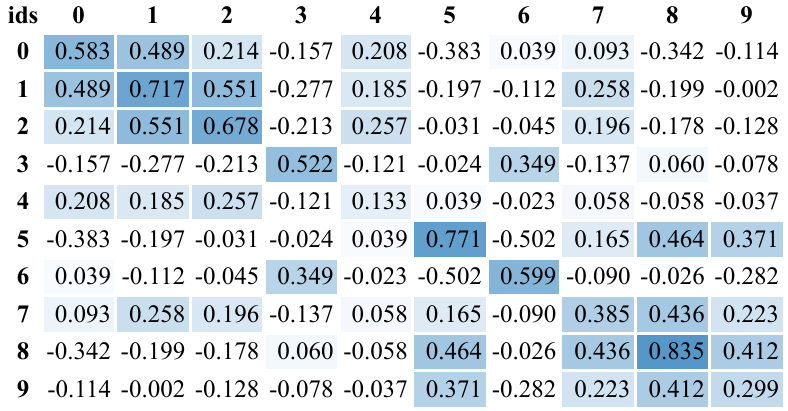}
    \rotatebox{90}{Shape}
    \centering
    \includegraphics[width=0.42\linewidth]{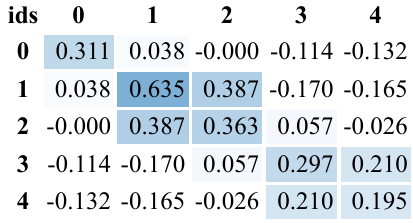}
    \rotatebox{90}{ Expression}
    \caption{Comparison of the shape parameters per shape instance (top) and expression parameters per expression instance (bottom) with an \textbf{uniformly sampled pose} as described in Section~\ref{sec:dataset}.}
    \label{fig:auto_cos_synth_pose}
    \end{subfigure}
    
    \begin{subfigure}[t,c]{\linewidth}
    \includegraphics[width=0.82\linewidth]{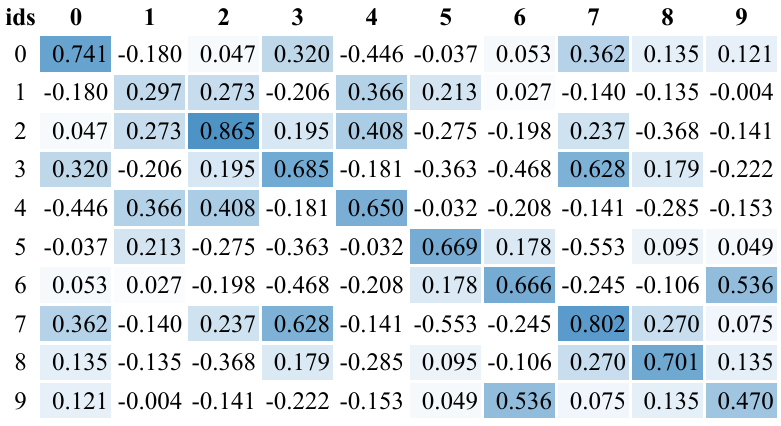}
    \rotatebox{90}{Shape}

    \centering
    \includegraphics[width=0.42\linewidth]{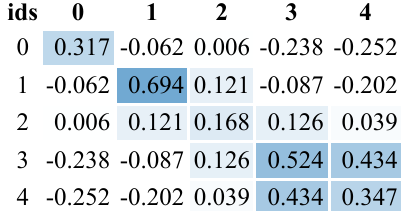}
    \rotatebox{90}{ Expression}
    \caption{Comparison of the shape parameters per shape instance (top) and expression parameters per expression instance (bottom) with a \textbf{neutral pose}.}
    \label{fig:auto_cos_synth_noPose}
    \end{subfigure}
    \caption{Cosine similarity comparison between the face model parameters computed by the autoencoder evaluated on \textit{synthetic} data input with a uniformly sampled pose (top) and a neutral pose (bottom). }
    \label{fig:auto_cos_synth}
\end{figure}

\begin{figure}
    \centering
    \begin{subfigure}[t,c]{\linewidth}
    \fontsize{7pt}{7pt}\selectfont
    \includegraphics[width=0.4\linewidth]{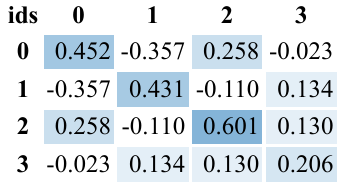}
    \rotatebox{90}{Shape}
    \includegraphics[width=0.43\linewidth]{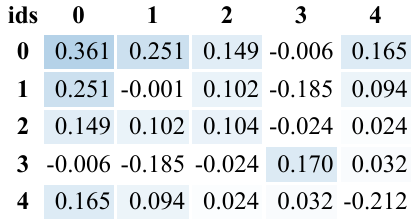}
    \rotatebox{90}{Expression}
      \caption{Comparison of the shape and expression parameters computed by the autoencoder evaluated on \textbf{real amplitude images}.}
      \label{fig:auto_cos_real_radar}
    \end{subfigure}

    \centering
    \begin{subfigure}[t,c]{\linewidth}
    \fontsize{7pt}{7pt}\selectfont
   \includegraphics[width=0.4\linewidth]{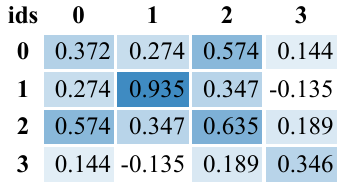}
   \rotatebox{90}{Shape}
     \includegraphics[width=0.43\linewidth]{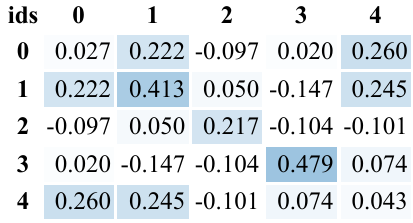}
     \rotatebox{90}{Expression}
      \caption{Comparison of the shape and expression parameters computed by the autoencoder evaluated on \textbf{real amplitude-depth images}.}
      \label{fig:auto_cos_real_combined}
    \end{subfigure}

    \centering
    \begin{subfigure}[t,c]{\linewidth}
    \fontsize{7pt}{7pt}\selectfont
    \includegraphics[width=0.4\linewidth]{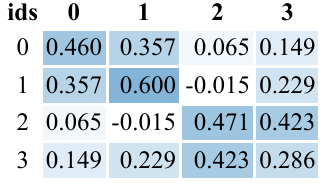}
    \rotatebox{90}{Shape}
     \includegraphics[width=0.43\linewidth]{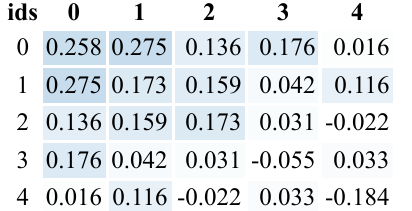}
     \rotatebox{90}{Expression}
      \caption{Comparison of the shape and expression parameters computed by the autoencoder evaluated on the \textbf{real-to-synth amplitude images}.}
      \label{fig:auto_cos_real_synth_radar}
    \end{subfigure}    
    \caption{Cosine similarity comparison between the shape and expression parameters from the autoencoder results derived from real data. The plots present the shape parameters (left) and expression parameters (right) grouped by instances of the same shape or expression, respectively.}
    \label{fig:auto_cos_real}
\end{figure}
\makeatletter
\begin{table*}
  \centering
  \begin{tabular}{p{0.005\linewidth} c c c c c c p{0.03\linewidth}}
    \vspace{-50pt} \rotatebox{90}{\centering Synthetic} & 
    \includegraphics[width=0.13\linewidth,trim={45px 45px 45px 45px},clip]{images/image_results/radar_images/synthetic/input/radar/8502_in_vir.png} 
    & \includegraphics[width=0.13\linewidth,trim={45px 45px 45px 45px},clip]{images/image_results/radar_images/synthetic/input/depth/8502_vir.png}
    & \includegraphics[width=0.13\linewidth,trim={45px 45px 45px 45px},clip]{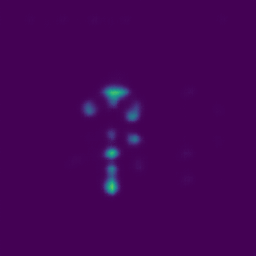} &  \includegraphics[width=0.13\linewidth,trim={45px 45px 45px 45px},clip]{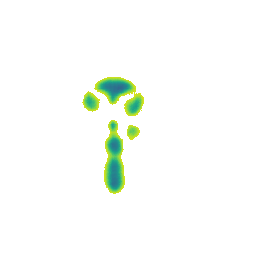} & \includegraphics[width=0.13\linewidth,trim={45px 45px 45px 45px},clip]{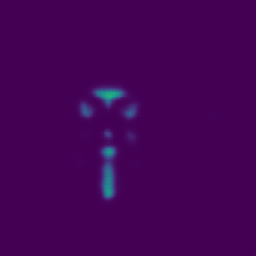} &
    \includegraphics[width=0.13\linewidth,trim={45px 45px 45px 45px},clip]{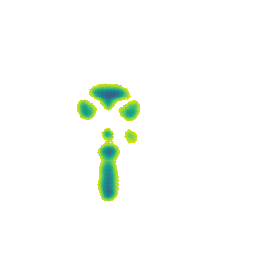} & \vspace{-40px}\hspace{-10px}\multirow{2}{*}{\includegraphics[width=\linewidth]{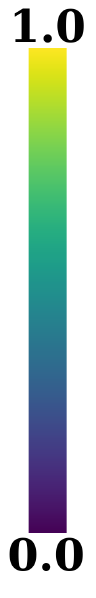}}
    \\
     \vspace{-40pt} \rotatebox{90}{\centering Real} & \includegraphics[width=0.13\linewidth,trim={45px 45px 45px 45px},clip]{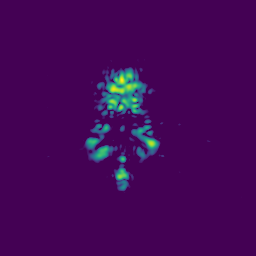}
    & \includegraphics[width=0.13\linewidth,trim={45px 45px 45px 45px},clip]{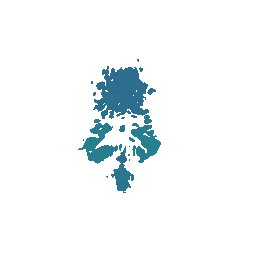}
    &\includegraphics[width=0.13\linewidth,trim={45px 45px 45px 45px},clip]{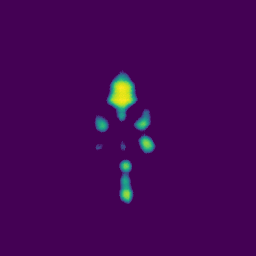} &  \includegraphics[width=0.13\linewidth,trim={45px 45px 45px 45px},clip]{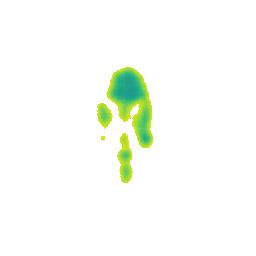} & \includegraphics[width=0.13\linewidth,trim={45px 45px 45px 45px},clip]{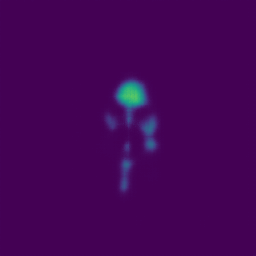} &
    \includegraphics[width=0.13\linewidth,trim={45px 45px 45px 45px},clip]{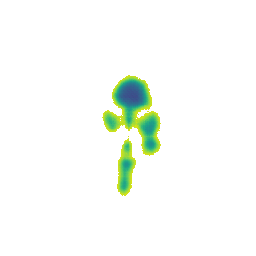} &\\
    & Amplitude Input & Depth Input & Amplitude Result & Depth Result & \multicolumn{2}{c}{Amplitude-Depth Result} &\\
  \end{tabular}
   \renewcommand{\@captype}{figure} 
  \caption{Reconstructions of the autoencoder models with the different settings. The first row contains the results for the synthetic radar image input for the differently trained models, while the second row contains the results for the real radar image input.}
  \label{fig:imgs_results}
\end{table*}
\makeatother

\makeatletter
\begin{table*}
    
\centering
\bgroup
\setlength\tabcolsep{0.1em}
\fontsize{7pt}{7pt}\selectfont
 \begin{tabular}{p{0.02\linewidth} p{0.10\linewidth} p{0\linewidth}}
            \vspace{-39pt}\multirow{2}{*}{\centering\rotatebox{90}{\centering \textbf{Synthetic}}} & \includegraphics[width=\linewidth]{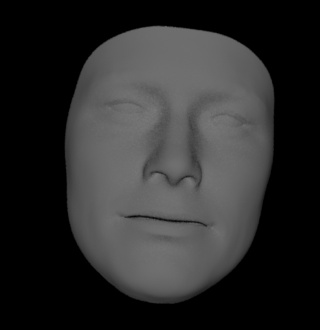} &\\
           & \centering Ground Truth &
         \end{tabular}
      \begin{tabular}{ p{0.11\linewidth} p{0.11\linewidth} p{0.11\linewidth} p{0.02\linewidth} p{0.11\linewidth} p{0.11\linewidth} p{0.11\linewidth} p{0.0\linewidth}}
        \centering
        \includegraphics[width=\linewidth]{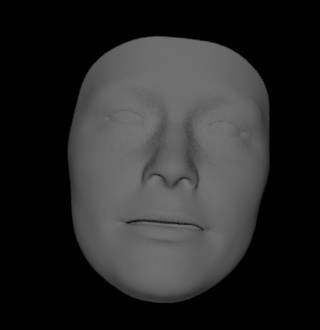} &  \includegraphics[width=\linewidth]{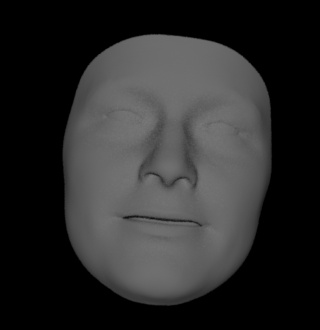} & \includegraphics[width=\linewidth]{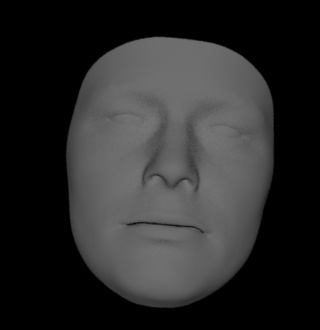} & 

        \centering \rotatebox{90}{\centering Encoder} &
        
        \includegraphics[width=\linewidth]{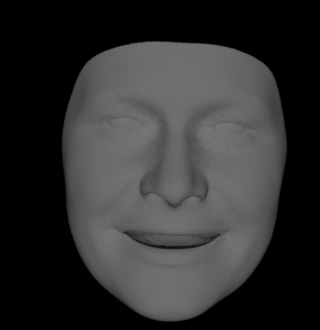} &  \includegraphics[width=\linewidth]{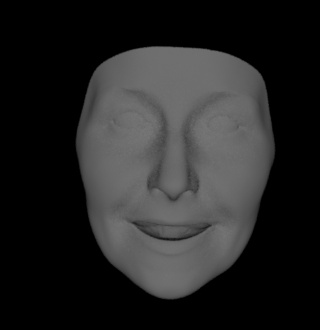} & \includegraphics[width=\linewidth]{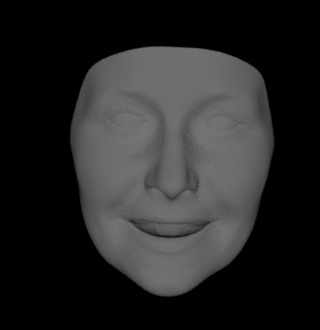} & \\
        
       \centering Amplitude  & \centering Depth & \centering Amplitude-Depth  & &  \centering Amplitude-Depth & \centering Depth  & \centering Amplitude &\\
        \includegraphics[width=\linewidth]{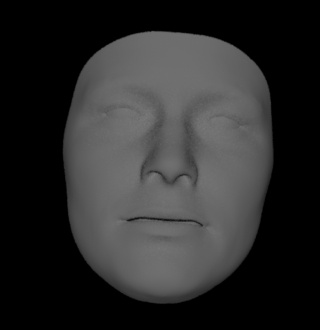} &  \includegraphics[width= \linewidth]{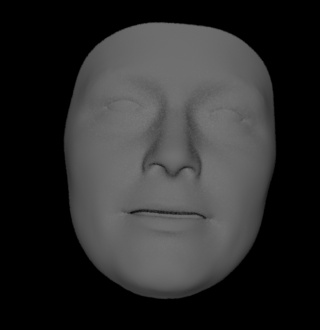} & \includegraphics[width=\linewidth]{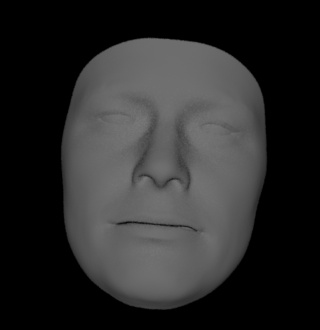} &

        \centering \rotatebox{90}{\centering Autoencoder} &
         
         \includegraphics[width=\linewidth]{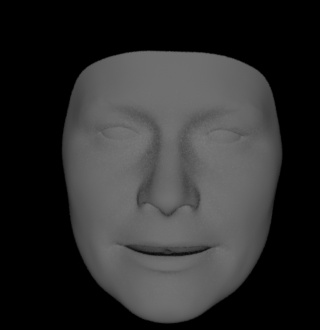} &  \includegraphics[width=\linewidth]{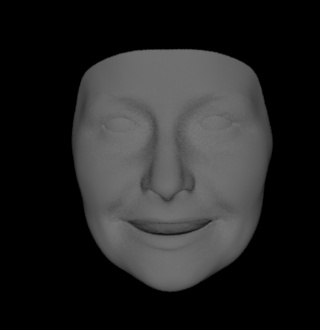} & \includegraphics[width=\linewidth]{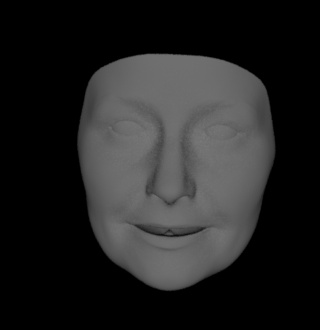} & \\
      \end{tabular}
       \begin{tabular}{p{0.11\linewidth} p{0.02\linewidth} p{0\linewidth}}
              \includegraphics[width=\linewidth]{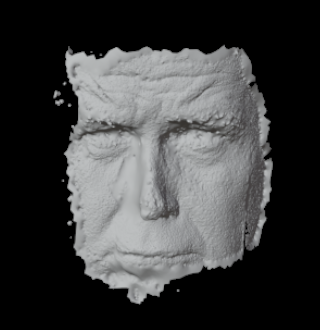} &
              \vspace{-35pt}\multirow{2}{*}{\centering\rotatebox{90}{\textbf{Real}}} & \\
            \centering Ground Truth & &
         \end{tabular}
\egroup
      \\
      \renewcommand{\@captype}{figure} 
  \caption{The resulting meshes generated with parameters of the different models. The left half depicts the results from the models evaluated on synthetic data, and the right half the results from the models evaluated on real data. }
  \label{fig:imgs_meshes}
\end{table*}
\makeatother

In the following the encoder and autoencoder architectures are evaluated quantitatively and qualitatively. 
The two network types are trained and evaluated using three types of radar input data: \textit{amplitude images}, \textit{depth images} that are reconstructed from radar signals, and the combination of both image types in two channels of an image (\textit{amplitude-depth images}). 
We further classify our data into three categories: synthetic images generated from synthetic meshes (\textit{synth data}), synthetic images from the photogrammetry reconstructions of real faces (\textit{real-to-synth data}), and real images from real faces (\textit{real data}). \\
\textbf{Quantitative Results}.
The L2 error between the predicted and ground truth face model parameters from the faces in our validation set is presented in Table~\ref{tab:loss_comparison}. Additionally, the mean Euclidean point distance between all corresponding mesh points of the face model meshes created with the result parameters are compared, without applying the pose. 
We compare these errors to the baseline, which is calculated by the error between the mean of the parameters within the training set and each face instance of the validation set. 
Furthermore, we compare the results to the error of randomly sampled parameters with the same distribution as the ground truth parameters as an additional baseline.  \\
All variants of our methods demonstrate superior outcomes compared to these baselines. 
The results of the encoder and autoencoder demonstrate that the L2 error of the resulting shape and expression vectors and, since the pose error is similar, the total error for the autoencoder is smaller than for the encoder. 
Furthermore, it can be observed that the results for the shape vector for both, the encoder and the autoencoder version, demonstrate a reduction in L2 error and the point distance of the corresponding meshes when the reconstructed depth image is incorporated. \\
In the following evaluations, we compare the shape and expression parameter results for \textit{real input}, \textit{real-to-synth input} and additionally generated \textit{synthetic input} which consists of shape instances having the same five sampled expressions. 
The sampling distributions are the same as in Section~\ref{sec:dataset}. 
The evaluation shows correlations between the input and result parameters, in particular for the real images. 
The Figures~\ref{fig:auto_cos_synth} and~\ref{fig:auto_cos_real} illustrate the comparison of the resulting faces with one another by computing the cosine similarity between the face model parameters and the parameters of the other face instances. %
Each cell contains the mean of face instances with the same shape or expression, respectively. 
Consequently, the similarity of faces can be compared to each other. The faces with the same ground truth shape, respectively expression, are anticipated to have the highest similarity. 
As a reference the cosine similarity of the ground truth parameters is presented in the Supplementary Material. %
Figure~\ref{fig:auto_cos_synth_pose} shows the comparison of the \textit{synthetic} shape vectors computed with an autoencoder trained on amplitude-depth images. 
The diagonal displays a high degree of similarity between vectors with the same ground truth shape, in comparison to other combinations of face vectors. 
While there are instances of two faces where the similarity between the shape vectors is also high, the values on the diagonal are the highest for most of the rows. 
Figure~\ref{fig:auto_cos_synth_noPose} illustrates the results for the same faces but without a pose. 
In this plot, the diagonal of high values is more prominent compared to the plot with a sampled pose. \\
The results for \textit{real} amplitude and amplitude-depth images to the autoencoder are displayed in Figures~\ref{fig:auto_cos_real_radar} and~\ref{fig:auto_cos_real_combined}. 
The shape results demonstrate higher values on the diagonal compared to the mean of the other values in the plot. 
With the exception of the first face in Figure~\ref{fig:auto_cos_real_combined}, these values represent the highest in each row/column. 
In Figure~\ref{fig:auto_cos_real_synth_radar} the results for the \textit{real-to-synth} version are evaluated, which also show the diagonal of high values for the shape evaluation. 
From the three variants only the autoencoder evaluated on real amplitude-depth images shows a diagonal in the comparison of the expression parameters. 
The results for the encoder and the other autoencoder versions can be found in the Supplementary Material. \\
To summarize, the comparison of the L2 loss shows that the autoencoder variants perform better in predicting the shape and expression parameters, while the use of the depth input improves the shape reconstruction. 
The cosine plots show that there is a correlation between the real face instance and the shape parameter output of the models, while there is no visible correlation for the expression vectors. 
We also showed that the pose influences the shape and expression parameter prediction. \\
\textbf{Qualitative Results.}
Qualitative results are presented in Figures~\ref{fig:imgs_results} and~\ref{fig:imgs_meshes}. 
For both, real and synthetic data, the image results appear to be a blurred version of the input, wherein smaller details, visible noise, and radar signal patterns have been removed. This is expected, as the physics-based renderer involves a random component. \\
Figure~\ref{fig:imgs_meshes} compares meshes created with the resulting parameters across the different input types. 
In the case of synthetic images, the results appear similar across the different methods. However, in the case of real input, the meshes exhibit greater variance and appear to deviate more from the ground truth. 
The predicted pose is found to be in close alignment with the ground truth pose for all methods. 

\section{Discussion}

\textbf{Evaluation.} The results of our experiments demonstrate that the autoencoder, as outlined in Section~\ref{sec:learned_renderer}, exhibits superior performance in predicting face model shape and expression parameters compared to the fully-supervised trained encoder, as described in Section~\ref{sec:encoder}. \\
We thus conclude that the additional training with the decoder has a beneficial effect on the training process, due to the image reconstruction task and its role in regularizing the loss of the network. 
Furthermore, the decoder component, which has been trained as a learned renderer, can be utilized to generate images with an appearance similar to radar images. 
Since it is differentiable and computes the images at a significantly faster rate compared to the physics-based renderer, it can further be utilized to finetune the parameters by fixing the encoder and decoder and optimize the image loss. \\
In terms of predicting the shape parameters, the depth images appear to demonstrate superior performance compared to the amplitude images, across both the autoencoder and encoder models. 
In particular, the prediction of the shape parameters is enhanced by the depth image variant. 
We assume that this is because the face pixel values are more evenly distributed and there are fewer small region peaks with high values, since this is the major difference between the image types. \\ 
\textbf{Limitations.} Despite the promising outcomes revealed by our evaluation, our work is not without limitations, thereby suggesting avenues for future research.
One challenge is the domain gap between synthetic and real images which decreases the resulting quality for real input. It appears in the different patterns between the radar simulator and the real radar images and different scales between the synthetic meshes and the real faces. Ideally, we could train our autoencoder on real images like Tewari et al.~\cite{tewari_mofa_2017}, but we are restricted to a limited real test set, which is also limited in variance since all radar captured persons are male and from a similar geographical region. \\ %
Another limitation is that the physics-based renderer only approximates the reflectance properties of skin, and therefore, our model also learns only an approximation, which is not physically correct. \\
\textbf{Future Work.} 
In future experiments, we propose collecting a larger dataset of real data incorporating a wider variety of human subjects for radar capture. \\
Another avenue is the evaluation of each viewing angle by the expected output quality of the face reconstruction, which can be insightful and useful in practice. 
This topic is already under investigation as current publications show~\cite{wirth2024maroonframeworkjointcharacterization}. \\
The method we present is generalizable to other applications than monitoring patients as long as it is a static environment and the person is within radar-capturing distance. 

\section{Conclusion}
We presented an approach for face reconstruction based on radar images and the BFM 2019~\cite{gerig_bfm_2017}. 
We generated a dataset of 10,000 synthetic radar images from a physics-based radar simulator based on meshes created with the BFM 2019 and trained three CNN models fully-supervised to reproduce the ground truth parameters. 
Furthermore, a learned renderer was employed, trained on rendering the radar images derived from the combination of the face model parameters and pose. 
This learned renderer is used to generate representative images close to the appearance of real data but significantly faster (more than 2000 times) and fully differentiable. 
While the reconstruction of faces only given radar data remains challenging, we demonstrated that the joint training process improves the reconstruction quality compared to the fully-supervised training approach. \\
On synthetic data, we achieve a mean Euclidean 3D point distance of 2.56~mm of the face meshes without applying the pose. Furthermore, the qualitative results appear visually similar to the ground truth faces. 
In the case of real data, we show that we can perform face recognition since instances of the same shape, but with a different pose and expression, have a higher similarity compared to other faces. However, the recognition of expressions is only possible with identity-specific trained models. 
The pose is consistent with the ground truth, however, for real data there is no discernible correlation between the ground truth and the resulting meshes concerning shape and expression in most of the results. 
The core benefit of our method is that it can be trained in an unsupervised fashion as model-based autoencoder on a large set of radar images without explicit 3D supervision which enables large scale training on real data. 
With our approach, we anticipate to guide future research towards higher-fidelity reconstructions.

\section{Acknowledgements}
This work was supported in part by Deutsche Forschungsgemeinschaft (DFG, German Research Foundation) – SFB 1483 – Project-ID 442419336, EmpkinS. The authors would like to thank the Rohde \& Schwarz GmbH \& Co. KG (Munich, Germany) for providing the radar imaging devices and technical support that made this study possible. We thank Nikolai Hofmann and Paul Himmler for proofreading the paper. The authors gratefully acknowledge the scientific support and HPC resources provided by the Erlangen National High Performance Computing Center of the Friedrich-Alexander-Universität Erlangen-Nürnberg.
{
    \small
    \bibliographystyle{ieeenat_fullname}
    \bibliography{main}
}
\clearpage
\setcounter{page}{1}
\setcounter{section}{0}
\maketitlesupplementary
\def\thesection{\Alph{section}}

\section{Additional Results}
\label{sec:cosine_sim}
\begin{figure}[h!]
    \centering
    \begin{subfigure}[t,c]{\linewidth}
     \includegraphics[width=0.94\linewidth]{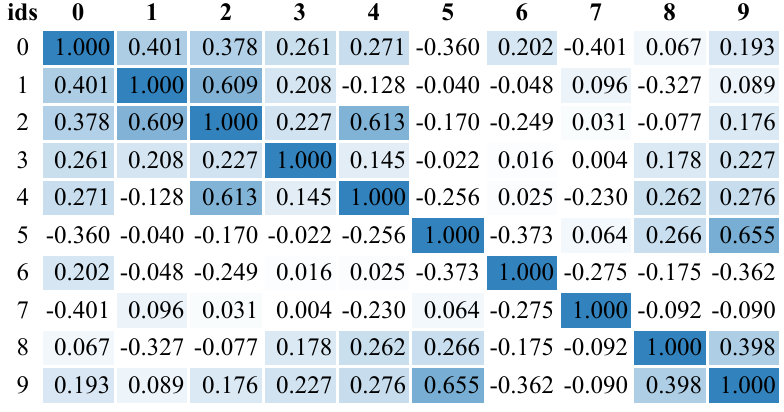}
     \rotatebox{90}{Shape}
    \caption{Shape parameters per shape instance.}
    \end{subfigure}
    \begin{subfigure}[t,c]{\linewidth}
    \centering
     \includegraphics[width=0.6\linewidth]{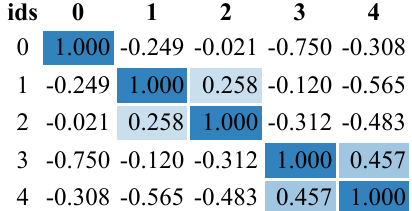}
     \rotatebox{90}{ Expression}
    \caption{Expression parameters per expression instance.}
    \end{subfigure}
   
    \caption{Comparison of the cosine similarity of the shape parameters per shape instance (top) and expression parameters per expression instance (bottom) from the \textbf{synthetic ground truth parameters}. The plot can be used as a reference to the results shown in Figure \ref{fig:auto_cos_synth}. }
    \label{fig:sup_auto_cos_input}
\end{figure}

\begin{figure}[h!]
    \centering
    \begin{subfigure}[t,c]{\linewidth}
     \includegraphics[width=0.94\linewidth]{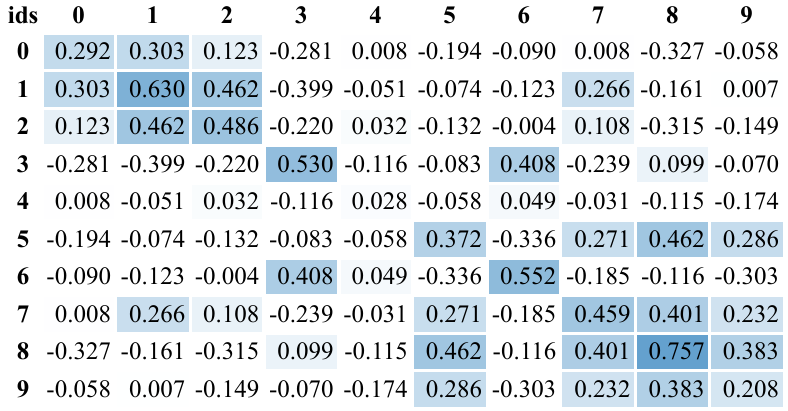}
     \rotatebox{90}{Shape}
    \caption{Shape parameters per shape instance.}
    \end{subfigure}
    \begin{subfigure}[t,c]{\linewidth}
    \centering
     \includegraphics[width=0.6\linewidth]{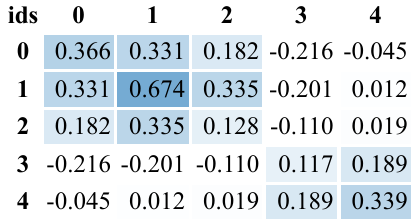}
     \rotatebox{90}{ Expression}
    \caption{Expression parameters per expression instance.}
    \end{subfigure}
   
    \caption{Comparison of the cosine similarity of the shape parameters per shape instance (top) and expression parameters per expression instance (bottom) from the \textbf{encoder} trained on synthetic \textbf{amplitude images}. }
    \label{fig:sup_enc_cos_amp}
\end{figure}

\begin{figure}[h!]
    \centering
    \begin{subfigure}[t,c]{\linewidth}
     \includegraphics[width=0.94\linewidth]{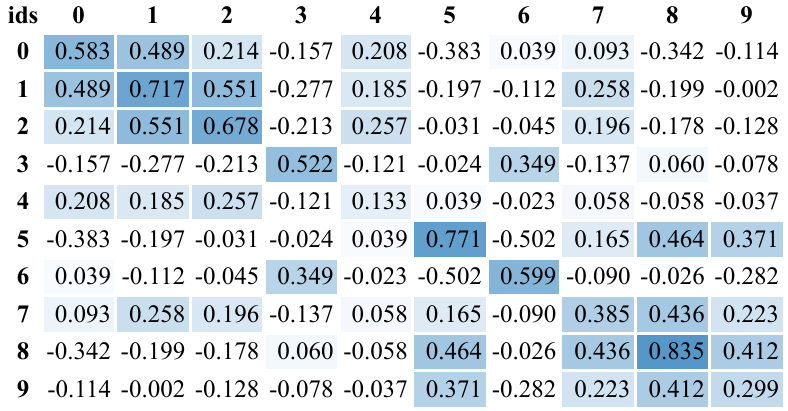}
     \rotatebox{90}{Shape}
    \caption{Shape parameters per shape instance.}
    \end{subfigure}
    \begin{subfigure}[t,c]{\linewidth}
    \centering
     \includegraphics[width=0.6\linewidth]{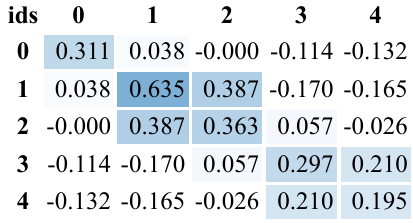}
     \rotatebox{90}{ Expression}
    \caption{Expression parameters per expression instance.}
    \end{subfigure}
   
    \caption{Comparison of the cosine similarity of the shape parameters per shape instance (top) and expression parameters per expression instance (bottom) from the \textbf{autoencoder} trained on synthetic \textbf{amplitude images}. }
    \label{fig:sup_auto_cos_amp}
\end{figure}

\begin{figure}[h!]
    \centering
    \begin{subfigure}[t,c]{\linewidth}
     \includegraphics[width=0.94\linewidth]{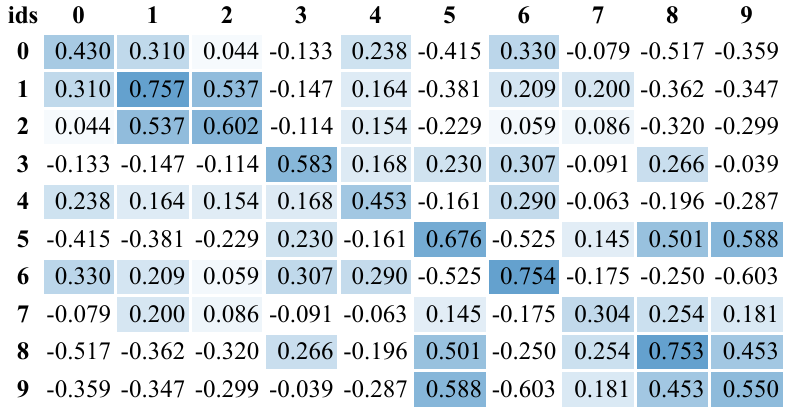}
     \rotatebox{90}{Shape}
    \caption{Shape parameters per shape instance.}
    \end{subfigure}
    \begin{subfigure}[t,c]{\linewidth}
    \centering
     \includegraphics[width=0.6\linewidth]{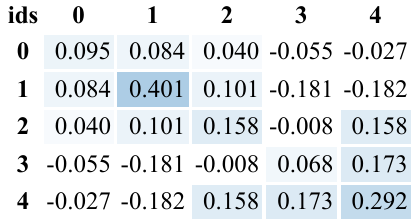}
     \rotatebox{90}{ Expression}
    \caption{Expression parameters per expression instance.}
    \end{subfigure}
   
    \caption{Comparison of the cosine similarity of the shape parameters per shape instance (top) and expression parameters per expression instance (bottom) from the \textbf{encoder} trained on synthetic \textbf{depth images}. }
    \label{fig:sup_enc_cos_depth}
\end{figure}

\begin{figure}[h!]
    \centering
    \begin{subfigure}[t,c]{\linewidth}
     \includegraphics[width=0.94\linewidth]{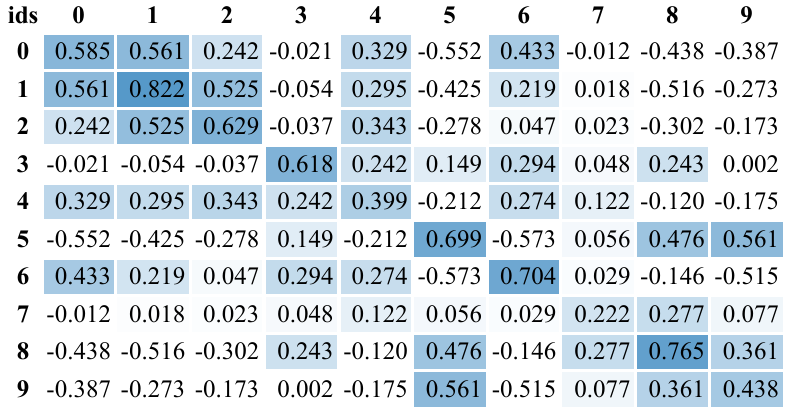}
     \rotatebox{90}{Shape}
    \caption{Shape parameters per shape instance.}
    \end{subfigure}
    \begin{subfigure}[t,c]{\linewidth}
    \centering
     \includegraphics[width=0.6\linewidth]{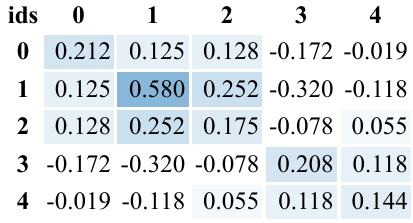}
     \rotatebox{90}{ Expression}
    \caption{Expression parameters per expression instance.}
    \end{subfigure}
   
    \caption{Comparison of the cosine similarity of the shape parameters per shape instance (top) and expression parameters per expression instance (bottom) from the \textbf{autoencoder} trained on synthetic \textbf{depth images}. }
    \label{fig:sup_auto_cos_depth}
\end{figure}

\begin{figure}[h!]
    \centering
    \begin{subfigure}[t,c]{\linewidth}
     \includegraphics[width=0.94\linewidth]{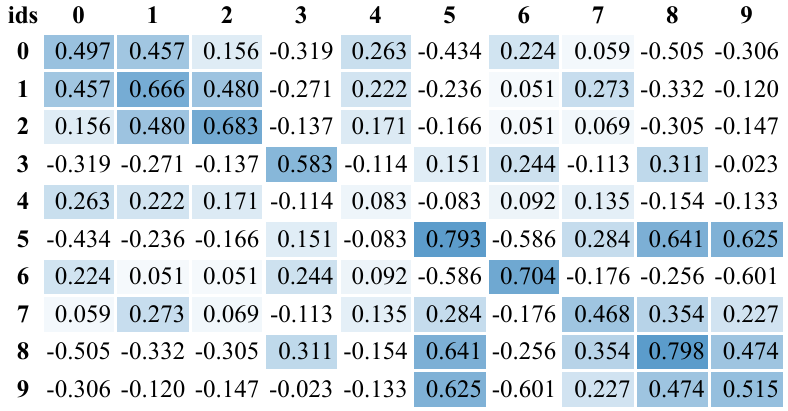}
     \rotatebox{90}{Shape}
    \caption{Shape parameters per shape instance.}
    \end{subfigure}
    \begin{subfigure}[t,c]{\linewidth}
    \centering
     \includegraphics[width=0.6\linewidth]{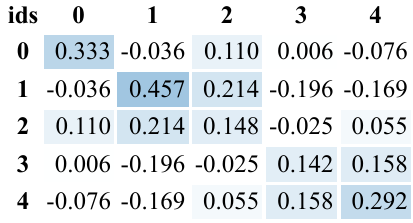}
     \rotatebox{90}{ Expression}
    \caption{Expression parameters per expression instance.}
    \end{subfigure}
   
    \caption{Comparison of the cosine similarity of the shape parameters per shape instance (top) and expression parameters per expression instance (bottom) from the \textbf{encoder} trained on synthetic \textbf{amplitude-depth images}. }
    \label{fig:sup_enc_cos_amp_depth}
\end{figure}

\begin{figure}[h!]
    \centering
    \begin{subfigure}[t,c]{\linewidth}
     \includegraphics[width=0.94\linewidth]{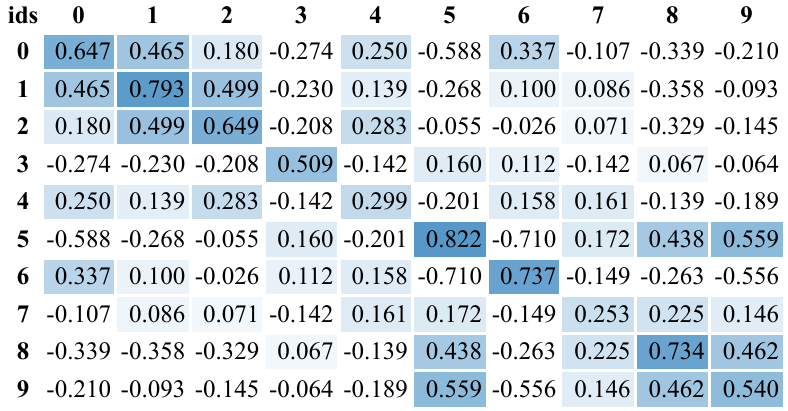}
     \rotatebox{90}{Shape}
    \caption{Shape parameters per shape instance.}
    \end{subfigure}
    \begin{subfigure}[t,c]{\linewidth}
    \centering
     \includegraphics[width=0.6\linewidth]{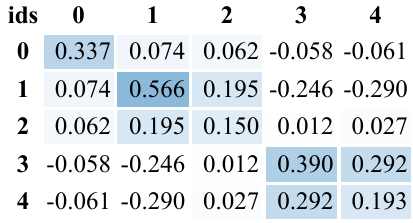}
     \rotatebox{90}{ Expression}
    \caption{Expression parameters per expression instance.}
    \end{subfigure}
   
    \caption{Comparison of the cosine similarity of the shape parameters per shape instance (top) and expression parameters per expression instance (bottom) from the \textbf{autoencoder} trained on synthetic \textbf{amplitude-depth images}. }
    \label{fig:sup_auto_cos_amp_depth}
\end{figure}

\begin{figure}[h!]
    \centering
    \begin{subfigure}[t,c]{\linewidth}
    \fontsize{7pt}{7pt}\selectfont
    \includegraphics[width=0.44\linewidth]{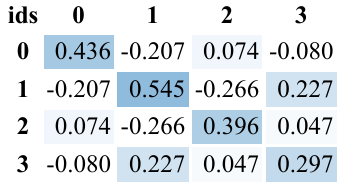}
    \rotatebox{90}{Shape}
    \includegraphics[width=0.49\linewidth]{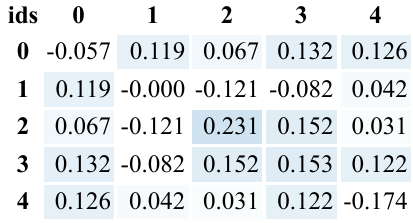}
    \rotatebox{90}{Expression}
      \caption{Comparison of the shape and expression parameters computed by the encoder evaluated on \textbf{real amplitude images}.}
    \end{subfigure}

    \centering
    \begin{subfigure}[t,c]{\linewidth}
    \fontsize{7pt}{7pt}\selectfont
   \includegraphics[width=0.44\linewidth]{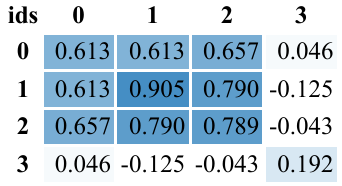}
   \rotatebox{90}{Shape}
     \includegraphics[width=0.49\linewidth]{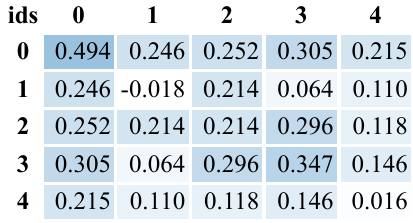}
     \rotatebox{90}{Expression}
      \caption{Comparison of the shape and expression parameters computed by the encoder evaluated on \textbf{real depth images}.}
    \end{subfigure}

    \centering
    \begin{subfigure}[t,c]{\linewidth}
    \fontsize{7pt}{7pt}\selectfont
    \includegraphics[width=0.44\linewidth]{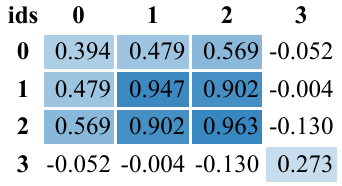}
    \rotatebox{90}{Shape}
     \includegraphics[width=0.49\linewidth]{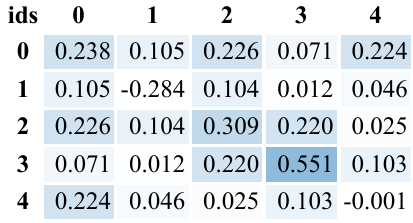}
     \rotatebox{90}{Expression}
      \caption{Comparison of the shape and expression parameters computed by the encoder evaluated on the \textbf{real amplitude-depth images}.}
    \end{subfigure}    
    \caption{Cosine similarity comparison between the shape and expression parameters from the \textbf{encoder} results derived from \textbf{real data}. The plots present the shape parameters (left) and expression parameters (right) grouped by instances of the same shape or expression, respectively.}
    \label{fig:sup_enc_cos_real}
\end{figure}

\begin{figure}[h!]
    \centering
    \begin{subfigure}[t,c]{\linewidth}
    \fontsize{7pt}{7pt}\selectfont
    \includegraphics[width=0.44\linewidth]{images/cosine_results/shape/real_radar_shape_crop.pdf}
    \rotatebox{90}{Shape}
    \includegraphics[width=0.49\linewidth]{images/cosine_results/expr/real_radar_expr_crop.pdf}
    \rotatebox{90}{Expression}
      \caption{Comparison of the shape and expression parameters computed by the autoencoder evaluated on \textbf{real amplitude images}.}
    \end{subfigure}

    \centering
    \begin{subfigure}[t,c]{\linewidth}
    \fontsize{7pt}{7pt}\selectfont
   \includegraphics[width=0.44\linewidth]{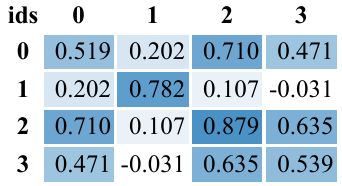}
   \rotatebox{90}{Shape}
     \includegraphics[width=0.49\linewidth]{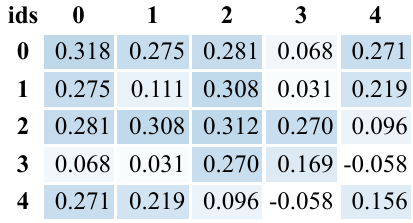}
     \rotatebox{90}{Expression}
      \caption{Comparison of the shape and expression parameters computed by the autoencoder evaluated on \textbf{real depth images}.}
    \end{subfigure}

    \centering
    \begin{subfigure}[t,c]{\linewidth}
    \fontsize{7pt}{7pt}\selectfont
    \includegraphics[width=0.44\linewidth]{images/cosine_results/shape/real_combined_shape_crop.pdf}
    \rotatebox{90}{Shape}
     \includegraphics[width=0.49\linewidth]{images/cosine_results/expr/real_combined_expr_crop.pdf}
     \rotatebox{90}{Expression}
      \caption{Comparison of the shape and expression parameters computed by the autoencoder evaluated on the \textbf{real amplitude-depth images}.}
    \end{subfigure}    
    \caption{Cosine similarity comparison between the shape and expression parameters from the \textbf{autoencoder} results derived from \textbf{real data}. The plots present the shape parameters (left) and expression parameters (right) grouped by instances of the same shape or expression, respectively.}
    \label{fig:sup_auto_cos_real}
\end{figure}

\textbf{Additional Quantitative Results.} In addition to the cosine similarity plots in Section~\ref{sec:results} of the paper, we provide additional comparisons of the ground truth parameters, the encoder results, and the autoencoder results evaluated on synthetic data. Figure~\ref{fig:sup_auto_cos_input} shows the ground truth parameters in comparison and can, therefore, be used as best case comparison for the other cosine similarity plots. The Figures~\ref{fig:sup_enc_cos_amp}, \ref{fig:sup_enc_cos_depth}, and \ref{fig:sup_enc_cos_amp_depth} show the cosine similarity of the resulting parameters from the encoder, and Figures~\ref{fig:sup_auto_cos_amp}, \ref{fig:sup_auto_cos_depth}, and \ref{fig:sup_auto_cos_amp_depth} the same for the autoencoder results. \\
In Figure~\ref{fig:sup_enc_cos_real} we compare the results from the encoder models evaluated on real data and in Figure~\ref{fig:sup_auto_cos_real} for the results from the autoencoder models. \\
\textbf{Additional Qualitative Results.} We also provide additional qualitative results of our method. The Figures~\ref{fig:sup_reco_image_synth} and \ref{fig:sup_reco_image_real} display several image results for the autoencoder trained and evaluated on amplitude, depth and amplitude-depth images. In Figures~\ref{fig:sup_mesh_synth} and \ref{fig:sup_mesh_real} we show additional examples for the resulting meshes of the encoder and autoencoder models trained on amplitude, depth, and amplitude-depth images. \\

\clearpage
\makeatletter
\begin{table*}[h!]
\bgroup
\setlength\tabcolsep{0.1em}
  \centering
  \begin{tabular}{c c | c c c c}
    Input Amp. & Input Depth & Amplitude & Depth & \multicolumn{2}{c}{Amp.-Depth}
    \\
    \midrule
    \includegraphics[width=0.16\linewidth]{images/image_results/radar_images/synthetic/input/radar/8502_in_vir.png} &
    \includegraphics[width=0.16\linewidth]{images/image_results/radar_images/synthetic/input/depth/8502_vir.png} &
    \includegraphics[width=0.16\linewidth]{images/image_results/radar_images/synthetic/autoencoder/radar/8502_vir.png} &
    \includegraphics[width=0.16\linewidth]{images/image_results/radar_images/synthetic/autoencoder/depth/8502_vir.png} &
    \includegraphics[width=0.16\linewidth]{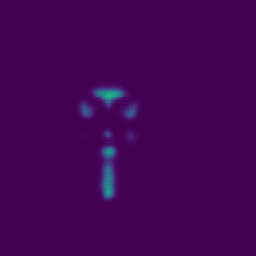} &
    \includegraphics[width=0.16\linewidth]{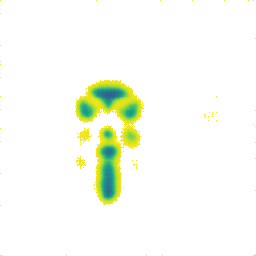} \\
    
    \includegraphics[width=0.16\linewidth]{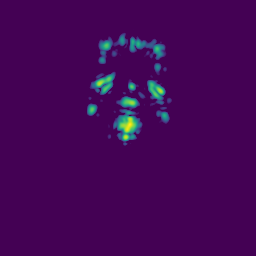} &
    \includegraphics[width=0.16\linewidth]{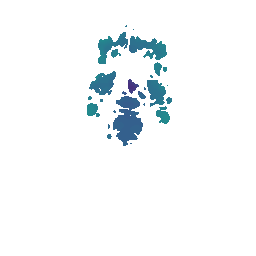} &
    \includegraphics[width=0.16\linewidth]{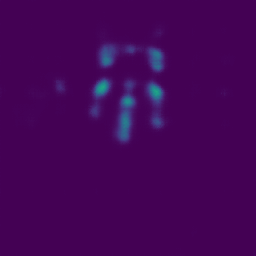} &
    \includegraphics[width=0.16\linewidth]{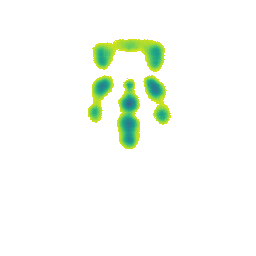} &
    \includegraphics[width=0.16\linewidth]{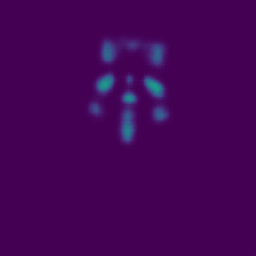} &
    \includegraphics[width=0.16\linewidth]{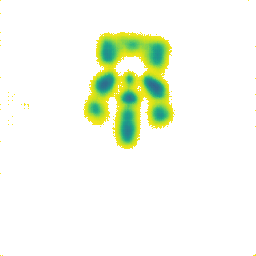}
    \\

    \includegraphics[width=0.16\linewidth]{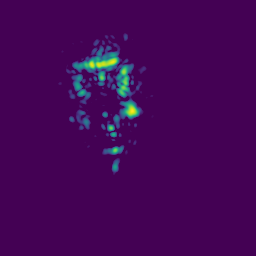} &
    \includegraphics[width=0.16\linewidth]{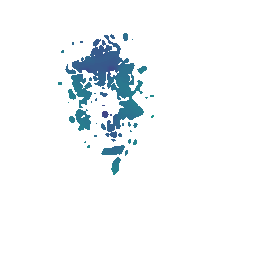} &
    \includegraphics[width=0.16\linewidth]{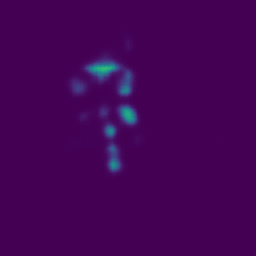} &
    \includegraphics[width=0.16\linewidth]{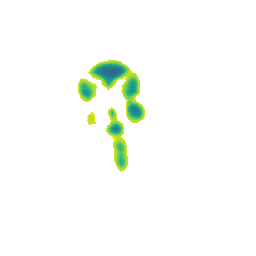} &
    \includegraphics[width=0.16\linewidth]{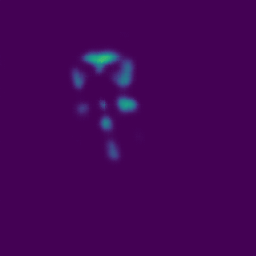} &
    \includegraphics[width=0.16\linewidth]{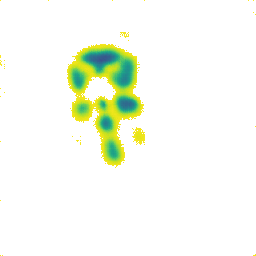}
    \\

    \includegraphics[width=0.16\linewidth]{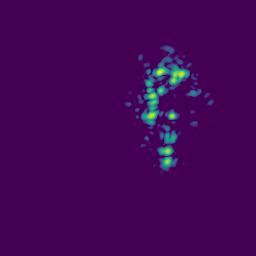} &
    \includegraphics[width=0.16\linewidth]{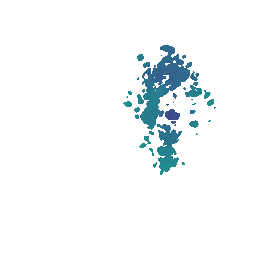} &
    \includegraphics[width=0.16\linewidth]{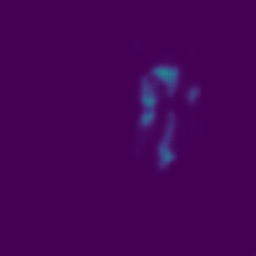} &
    \includegraphics[width=0.16\linewidth]{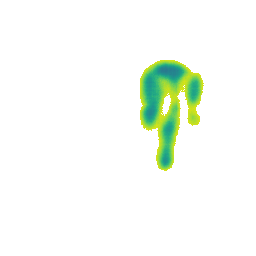} &
    \includegraphics[width=0.16\linewidth]{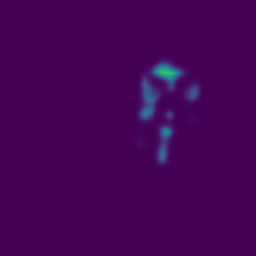} &
    \includegraphics[width=0.16\linewidth]{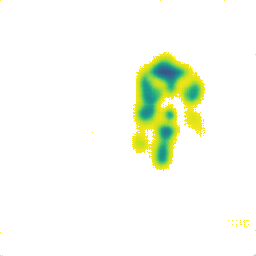}
    \\
    \includegraphics[width=0.16\linewidth]{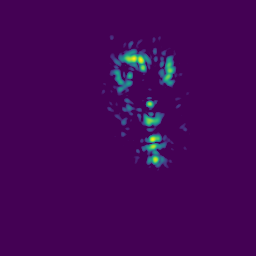} &
    \includegraphics[width=0.16\linewidth]{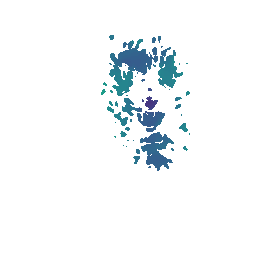} &
    \includegraphics[width=0.16\linewidth]{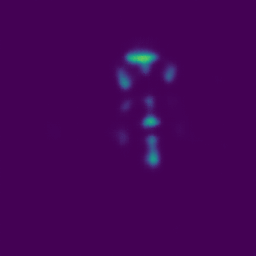} &
    \includegraphics[width=0.16\linewidth]{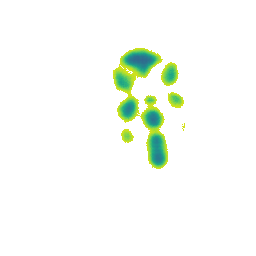} &
    \includegraphics[width=0.16\linewidth]{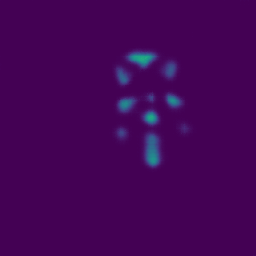} &
    \includegraphics[width=0.16\linewidth]{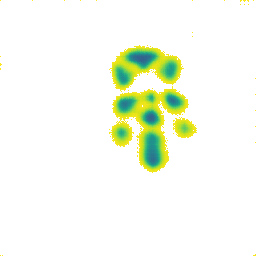}
    \\
  \end{tabular}
\egroup
   \renewcommand{\@captype}{figure} 
  \caption{Resulting amplitude images of the \textbf{autoencoder} model trained on \textbf{amplitude}, \textbf{depth}, and \textbf{amplitude-depth images} for different \textbf{synthetic} face instances. }
  \label{fig:sup_reco_image_synth}
\end{table*}
\makeatother
\makeatletter
\begin{table*}[h!]
\bgroup
\setlength\tabcolsep{0.1em}
  \centering
  \begin{tabular}{c c | c c c c}
    Input Amp. & Input Depth & Amplitude & Depth & \multicolumn{2}{c}{Amp.-Depth}
    \\
        \midrule
    \includegraphics[width=0.16\linewidth]{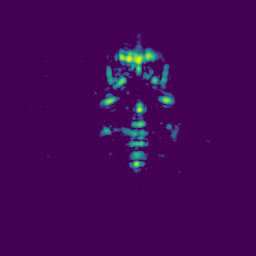} &
    \includegraphics[width=0.16\linewidth]{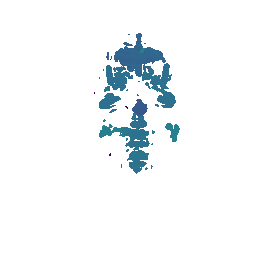} &
    \includegraphics[width=0.16\linewidth]{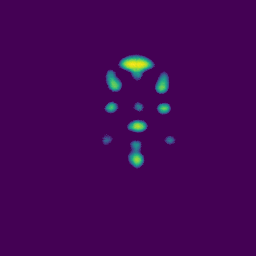} &
    \includegraphics[width=0.16\linewidth]{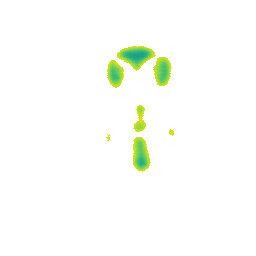} &
    \includegraphics[width=0.16\linewidth]{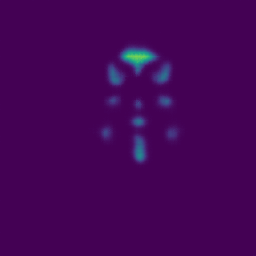} &
    \includegraphics[width=0.16\linewidth]{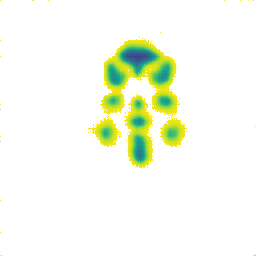} \\
    
    \includegraphics[width=0.16\linewidth]{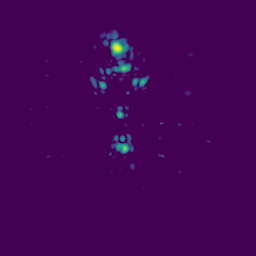} &
    \includegraphics[width=0.16\linewidth]{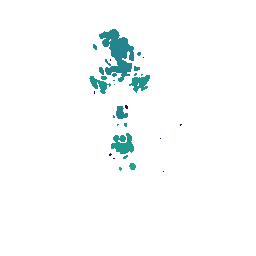} &
    \includegraphics[width=0.16\linewidth]{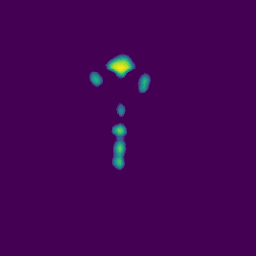} &
    \includegraphics[width=0.16\linewidth]{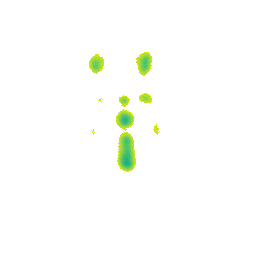} &
    \includegraphics[width=0.16\linewidth]{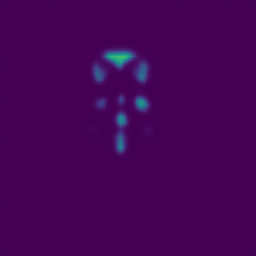} &
    \includegraphics[width=0.16\linewidth]{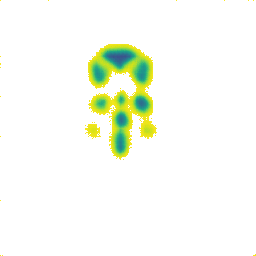}
    \\

    \includegraphics[width=0.16\linewidth]{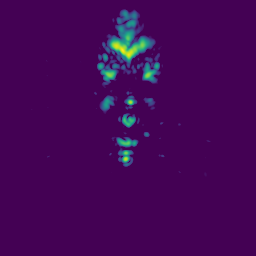} &
    \includegraphics[width=0.16\linewidth]{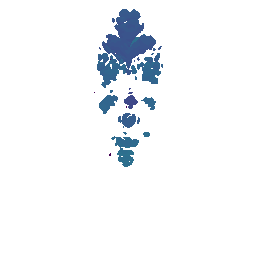} &
    \includegraphics[width=0.16\linewidth]{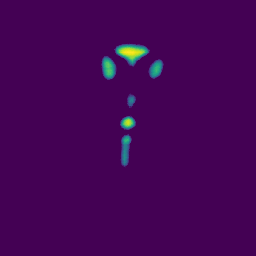} &
    \includegraphics[width=0.16\linewidth]{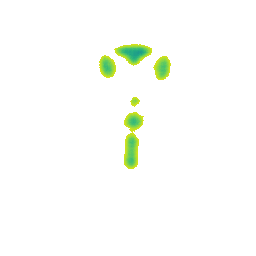} &
    \includegraphics[width=0.16\linewidth]{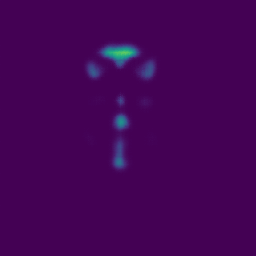} &
    \includegraphics[width=0.16\linewidth]{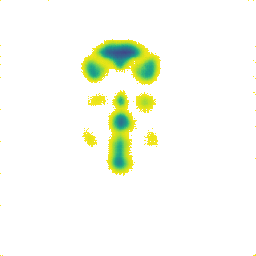}
    \\

    \includegraphics[width=0.16\linewidth]{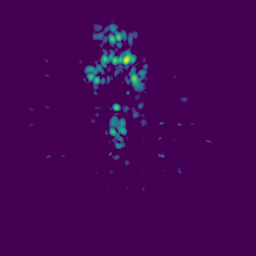} &
    \includegraphics[width=0.16\linewidth]{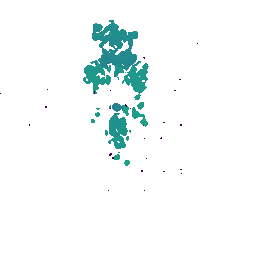} &
    \includegraphics[width=0.16\linewidth]{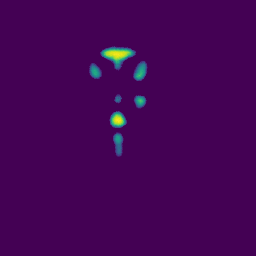} &
    \includegraphics[width=0.16\linewidth]{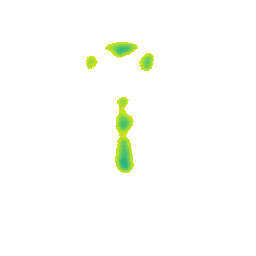} &
    \includegraphics[width=0.16\linewidth]{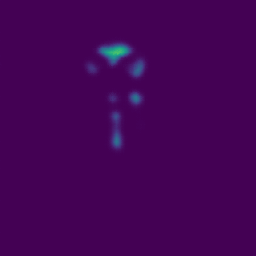} &
    \includegraphics[width=0.16\linewidth]{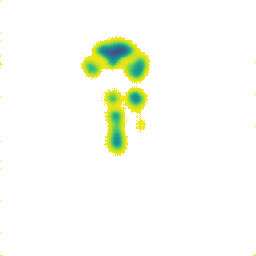}
    \\
    \includegraphics[width=0.16\linewidth]{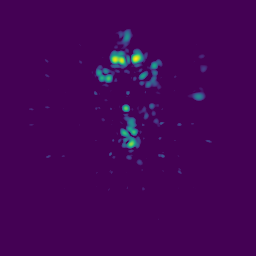} &
    \includegraphics[width=0.16\linewidth]{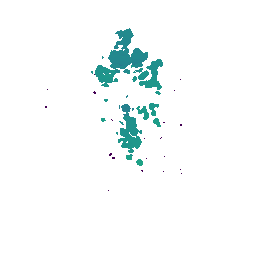} &
    \includegraphics[width=0.16\linewidth]{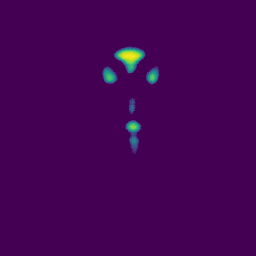} &
    \includegraphics[width=0.16\linewidth]{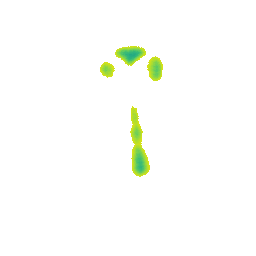} &
    \includegraphics[width=0.16\linewidth]{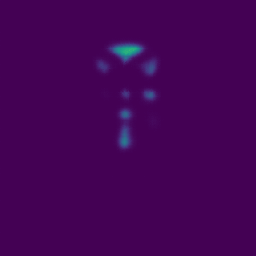} &
    \includegraphics[width=0.16\linewidth]{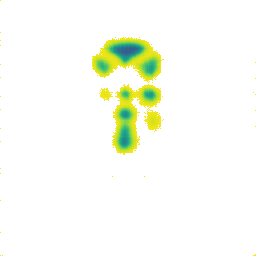}
    \\
  \end{tabular}
\egroup
   \renewcommand{\@captype}{figure} 
  \caption{Resulting amplitude images of the \textbf{autoencoder} model trained on \textbf{amplitude}, \textbf{depth}, and \textbf{amplitude-depth images} for different \textbf{real} face instances. }
  \label{fig:sup_reco_image_real}
\end{table*}
\makeatother
\makeatletter
\begin{figure*}[h!]
  \centering
  \begin{tabular}{p{0.015\linewidth} p{0.015\linewidth} p{0.9\linewidth}}
    & 
    \begin{minipage}{\linewidth}
    \rotatebox{90}{Input}
    \end{minipage} & 
    \begin{minipage}{\linewidth}
    \begin{subfigure}{\linewidth}
        \includegraphics[width=0.19\linewidth,trim={130px 50px 130px 50px},clip]{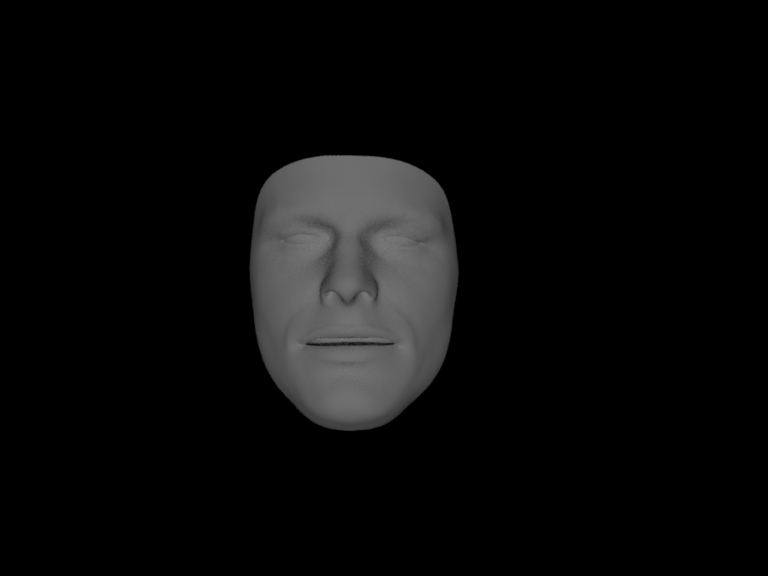} 
        \includegraphics[width=0.19\linewidth,trim={130px 100px 130px 0px},clip]{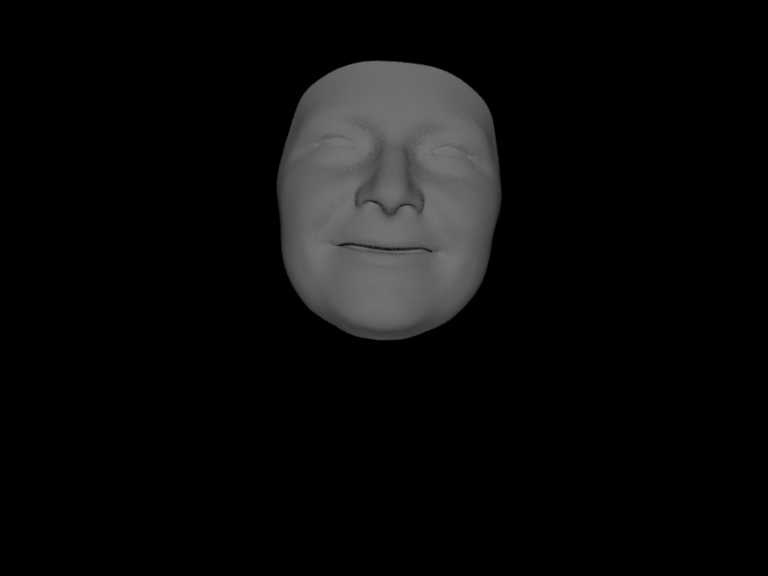} 
        \includegraphics[width=0.19\linewidth,trim={130px 50px 130px 50px},clip]{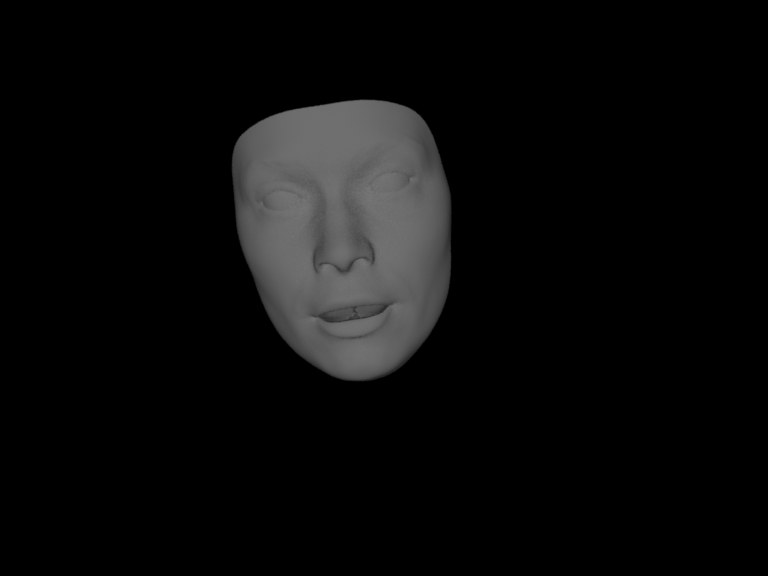} 
        \includegraphics[width=0.19\linewidth,trim={190px 50px 70px 50px},clip]{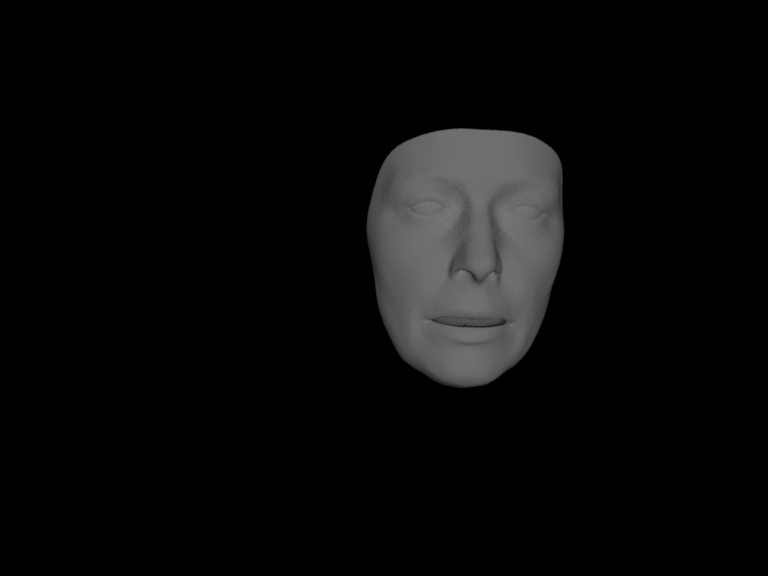} 
        \includegraphics[width=0.19\linewidth,trim={130px 50px 130px 50px},clip]{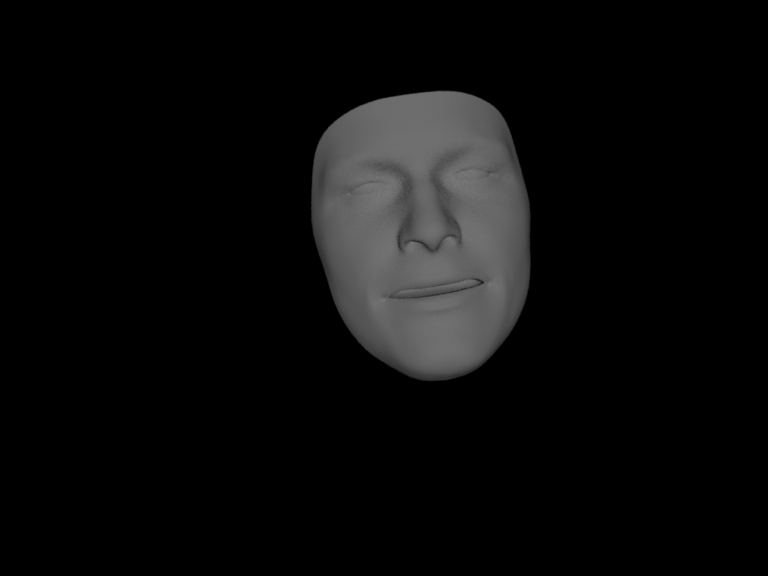} 
    \end{subfigure}
    \end{minipage} 
    \\
    \midrule

      & 
    \begin{minipage}{\linewidth}
    \rotatebox{90}{Amplitude}
    \end{minipage}  & 
    \begin{minipage}{\linewidth}
    \begin{subfigure}{\linewidth}
    \includegraphics[width=0.19\linewidth,trim={130px 50px 130px 50px},clip]{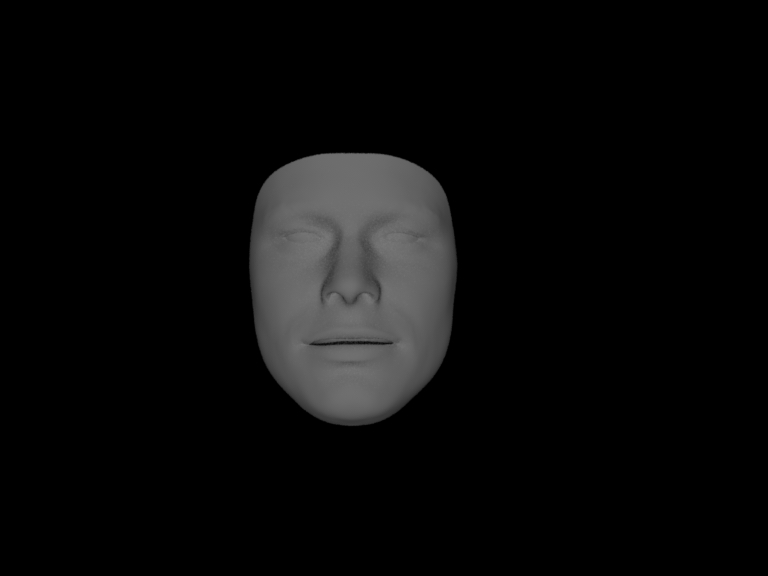} 
    \includegraphics[width=0.19\linewidth,trim={130px 100px 130px 0px},clip]{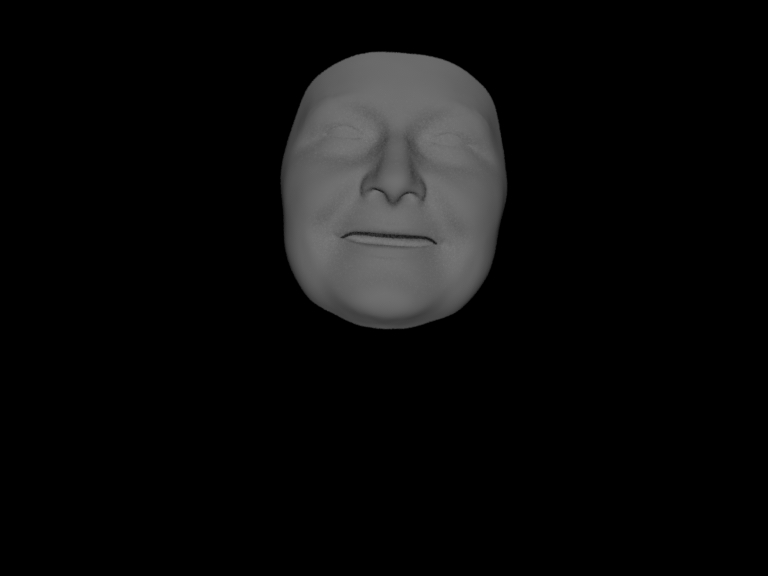} 
    \includegraphics[width=0.19\linewidth,trim={130px 50px 130px 50px},clip]{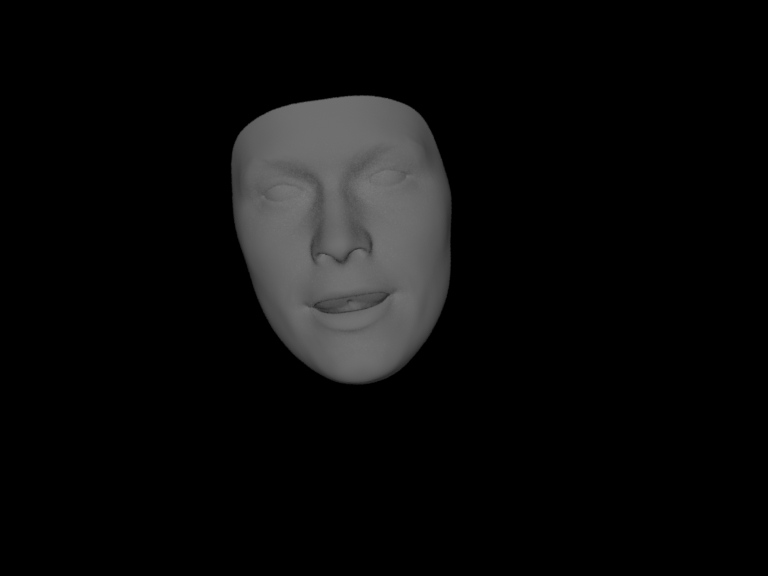} 
    \includegraphics[width=0.19\linewidth,trim={190 50px 70px 50px},clip]{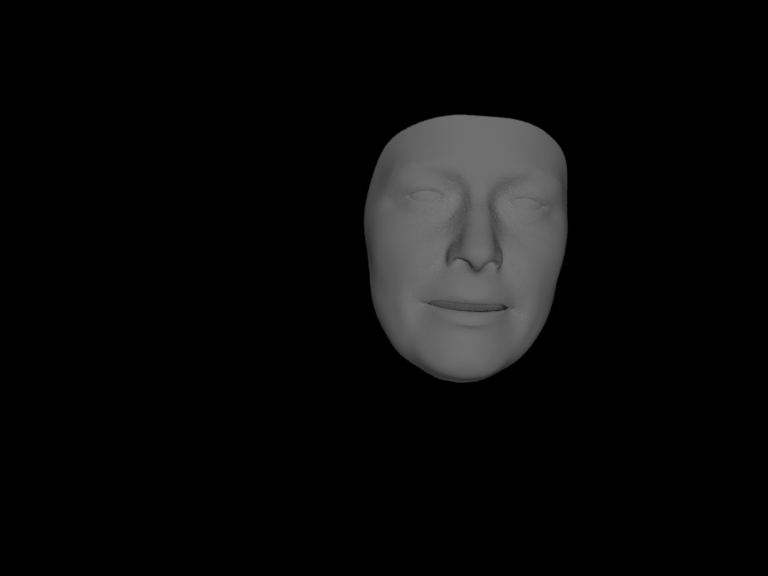} 
    \includegraphics[width=0.19\linewidth,trim={130px 50px 130px 50px},clip]{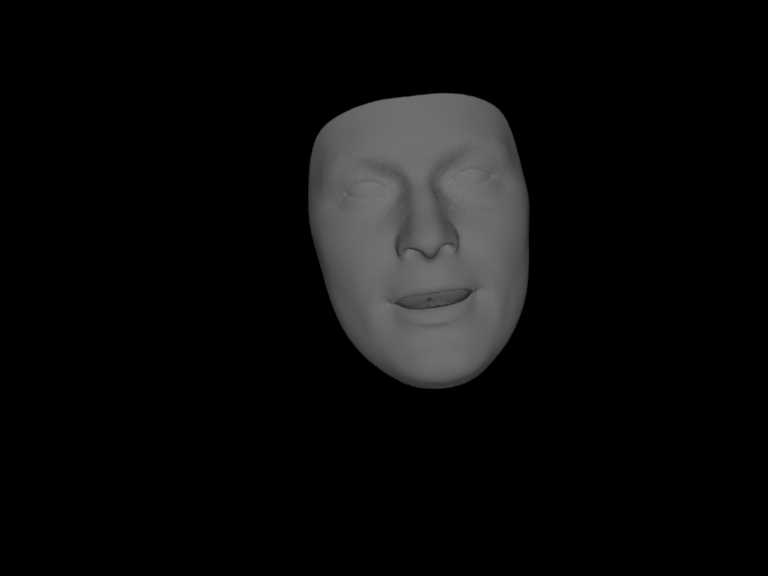}
    \end{subfigure}
    \end{minipage}
    \\
    \begin{minipage}{\linewidth}\rotatebox{90}{\textbf{Encoder}}\end{minipage}
    & \begin{minipage}{\linewidth}
    \rotatebox{90}{Depth}
    \end{minipage} & 
    \begin{minipage}{\linewidth}
    \begin{subfigure}{\linewidth}
    \includegraphics[width=0.19\linewidth,trim={130px 50px 130px 50px},clip]{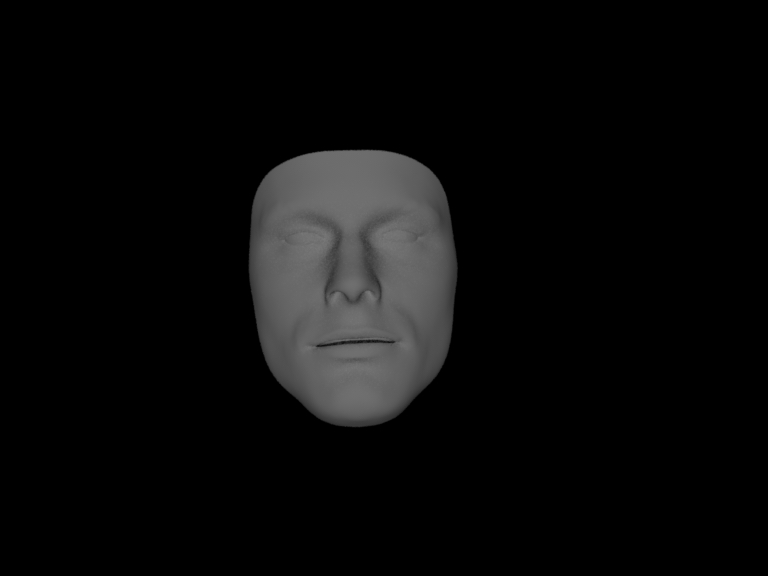} 
    \includegraphics[width=0.19\linewidth,trim={130px 100px 130px 0px},clip]{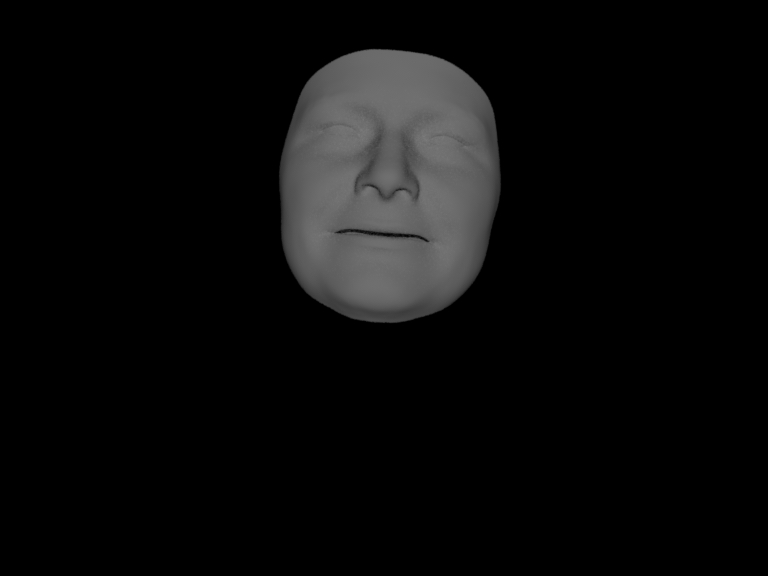} 
    \includegraphics[width=0.19\linewidth,trim={130px 50px 130px 50px},clip]{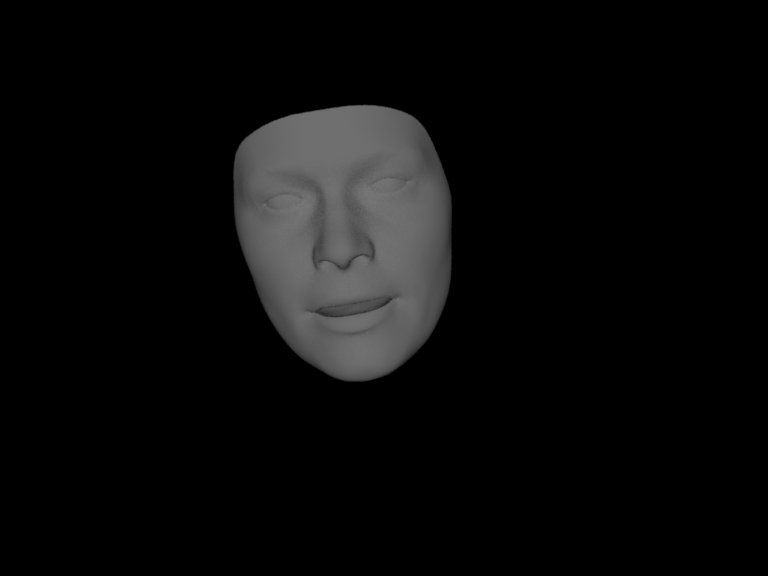} 
    \includegraphics[width=0.19\linewidth,trim={190 50px 70px 50px},clip]{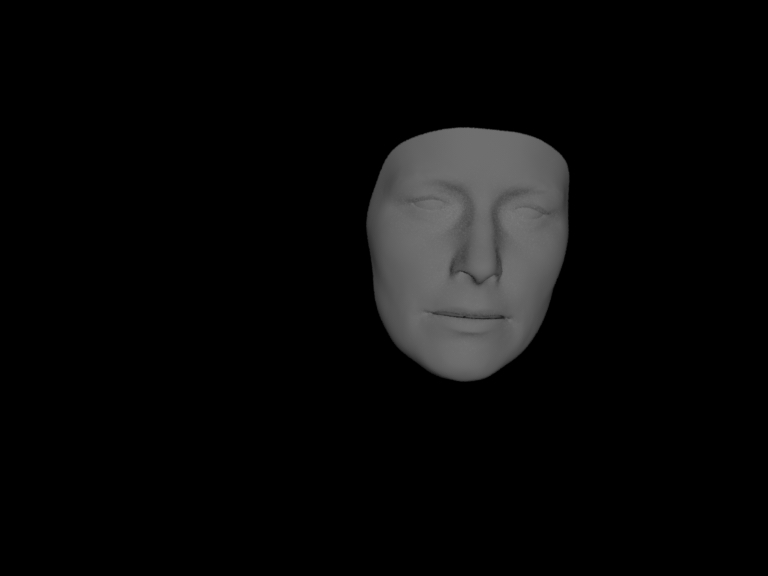} 
    \includegraphics[width=0.19\linewidth,trim={130px 50px 130px 50px},clip]{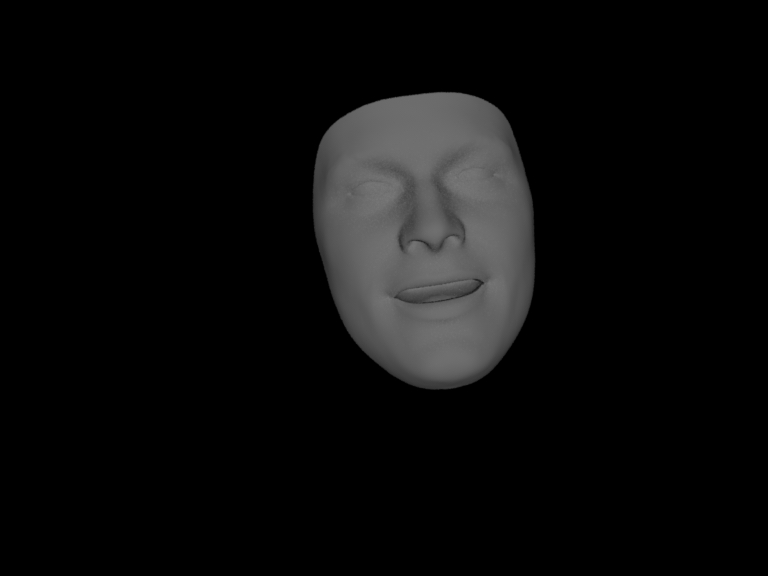}
    \end{subfigure}
    \end{minipage}
    \\
    & \begin{minipage}{\linewidth}
    \rotatebox{90}{Amp.-Depth}
    \end{minipage} & 
        \begin{minipage}{\linewidth}
    \begin{subfigure}{\linewidth}
    \includegraphics[width=0.19\linewidth,trim={130px 50px 130px 50px},clip]{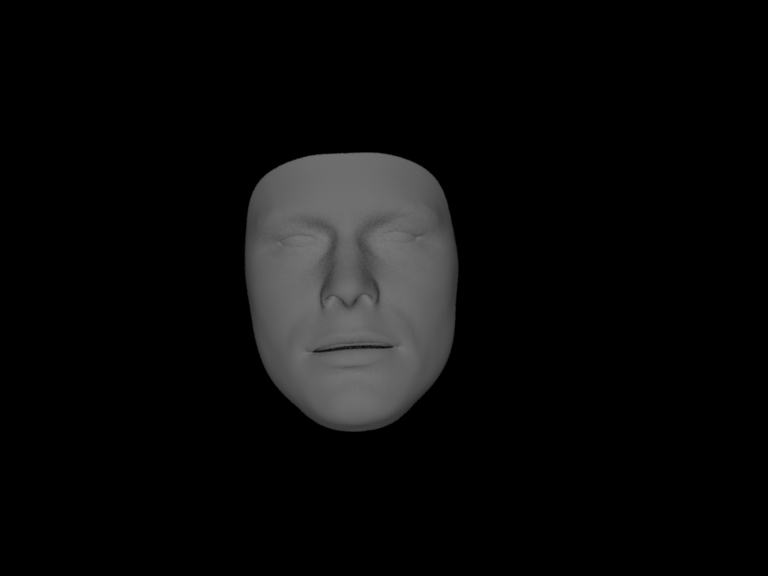}  
    \includegraphics[width=0.19\linewidth,trim={130px 100px 130px 0px},clip]{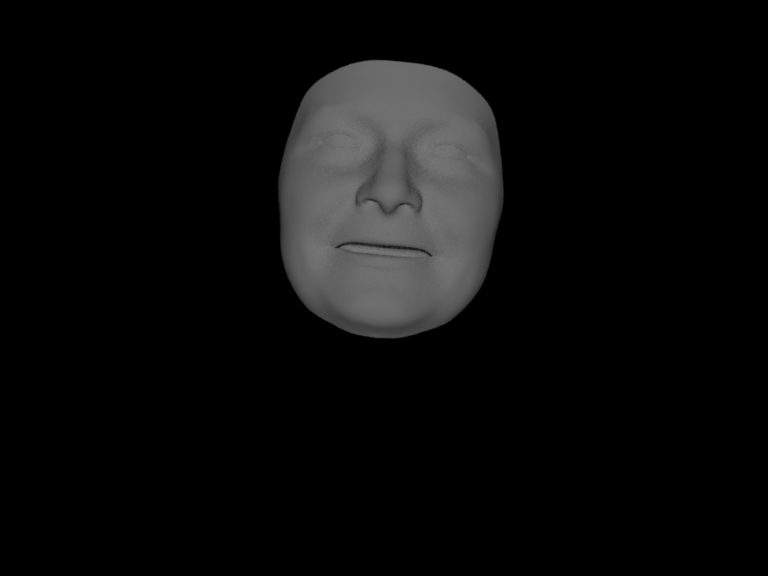}  
    \includegraphics[width=0.19\linewidth,trim={130px 50px 130px 50px},clip]{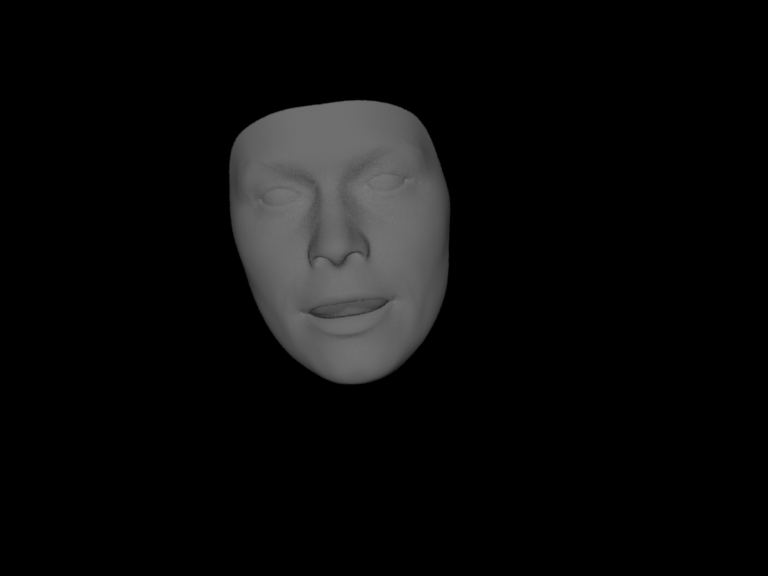}  
    \includegraphics[width=0.19\linewidth,trim={190 50px 70px 50px},clip]{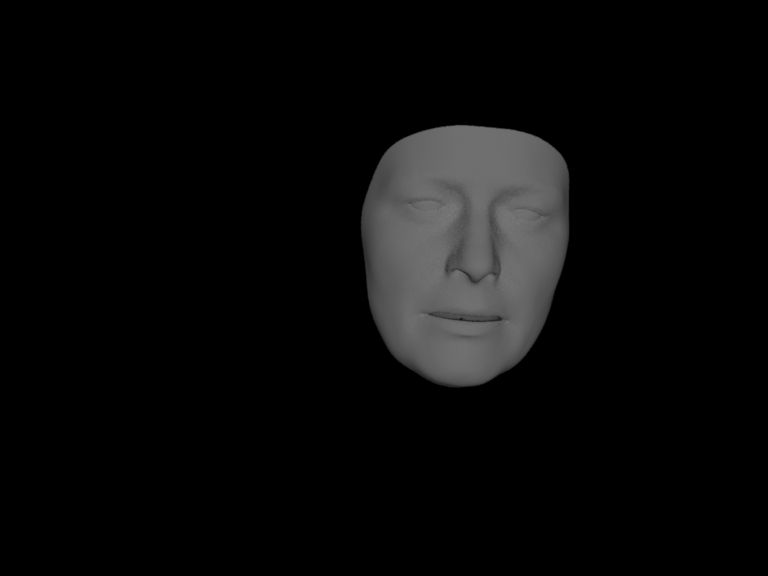}  
    \includegraphics[width=0.19\linewidth,trim={130px 50px 130px 50px},clip]{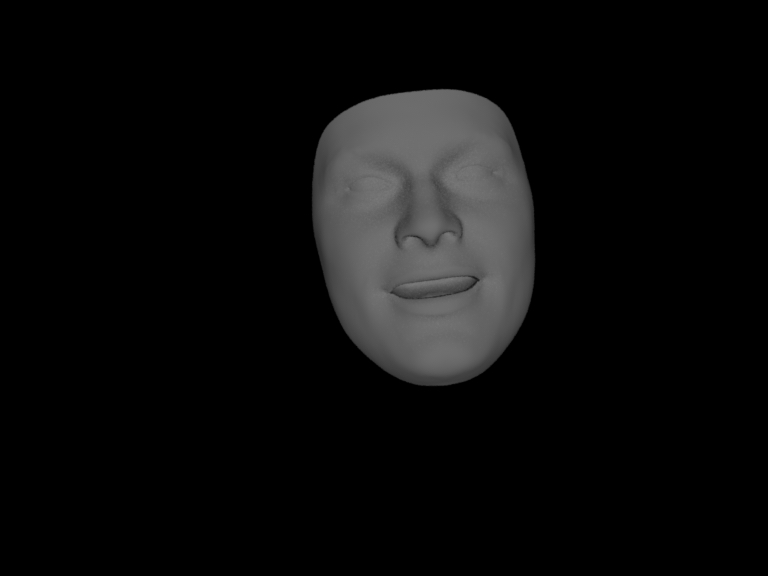}
    \end{subfigure}
    \end{minipage}
    \\
    \midrule
    & 
    \begin{minipage}{\linewidth}
    \rotatebox{90}{Amplitude}
    \end{minipage} & 
        \begin{minipage}{\linewidth}
    \begin{subfigure}{\linewidth}
    \includegraphics[width=0.19\linewidth,trim={130px 50px 130px 50px},clip]{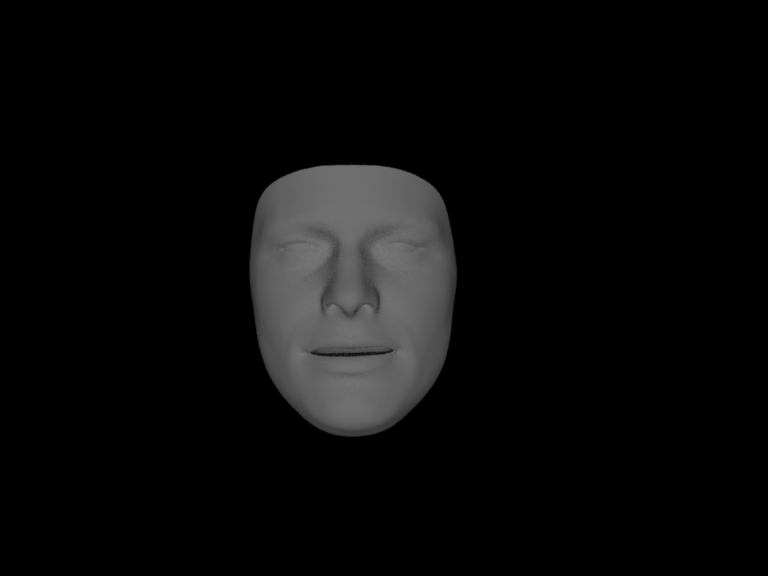} 
    \includegraphics[width=0.19\linewidth,trim={130px 100px 130px 0px},clip]{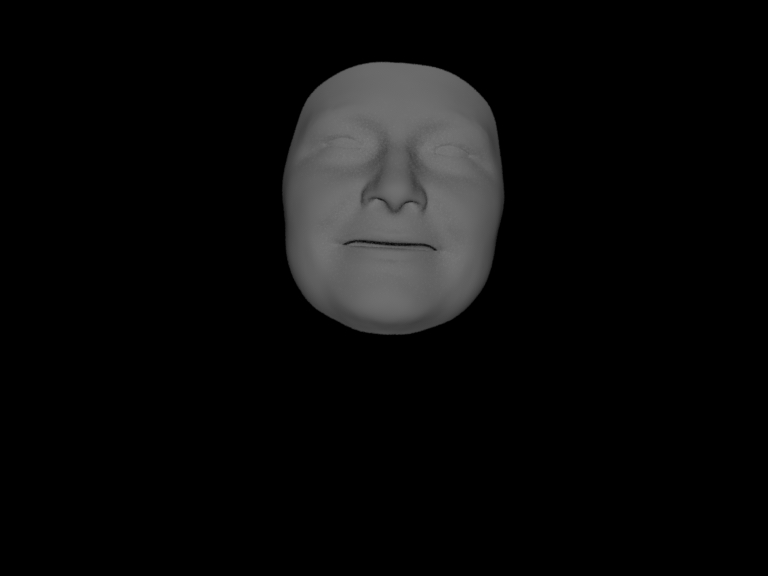} 
    \includegraphics[width=0.19\linewidth,trim={130px 50px 130px 50px},clip]{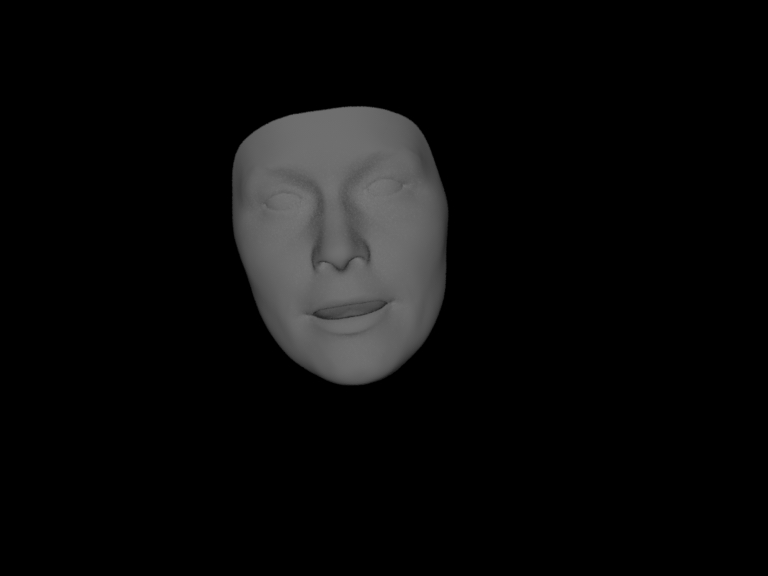}
    \includegraphics[width=0.19\linewidth,trim={190 50px 70px 50px},clip]{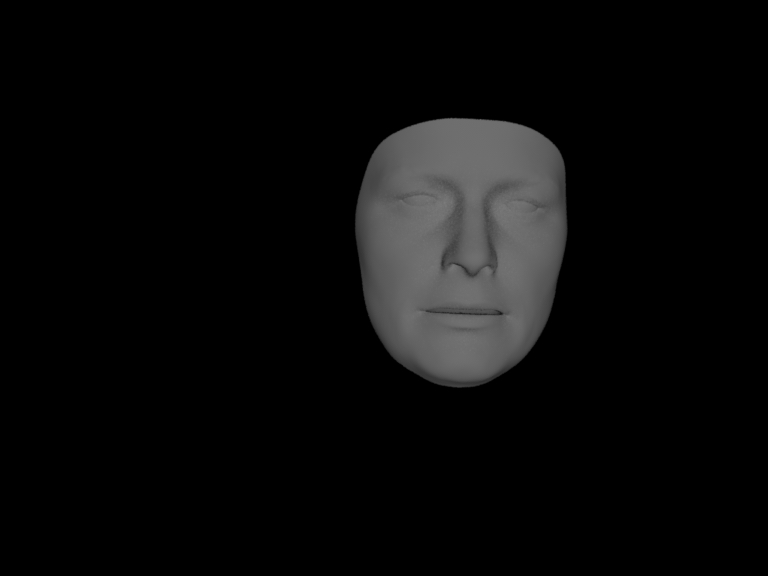} 
    \includegraphics[width=0.19\linewidth,trim={130px 50px 130px 50px},clip]{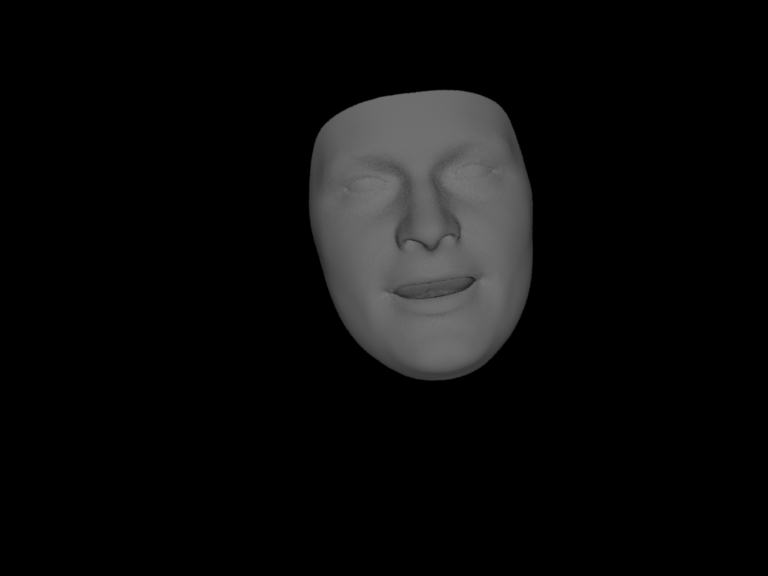}
    \end{subfigure}
    \end{minipage}
    \\
    \begin{minipage}{\linewidth}\rotatebox{90}{\textbf{Autoencoder}}\end{minipage} & \begin{minipage}{\linewidth}
    \rotatebox{90}{Depth}
    \end{minipage} & 
        \begin{minipage}{\linewidth}
    \begin{subfigure}{\linewidth}
    \includegraphics[width=0.19\linewidth,trim={130px 50px 130px 50px},clip]{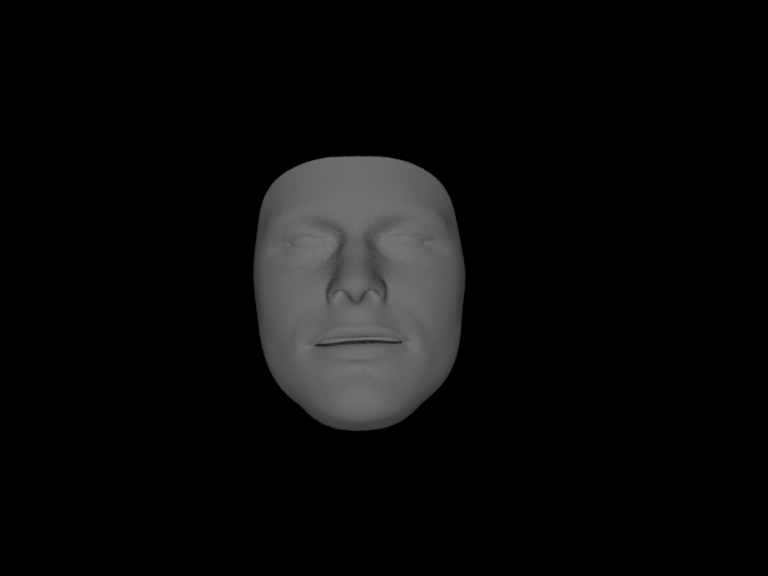} 
    \includegraphics[width=0.19\linewidth,trim={130px 100px 130px 0px},clip]{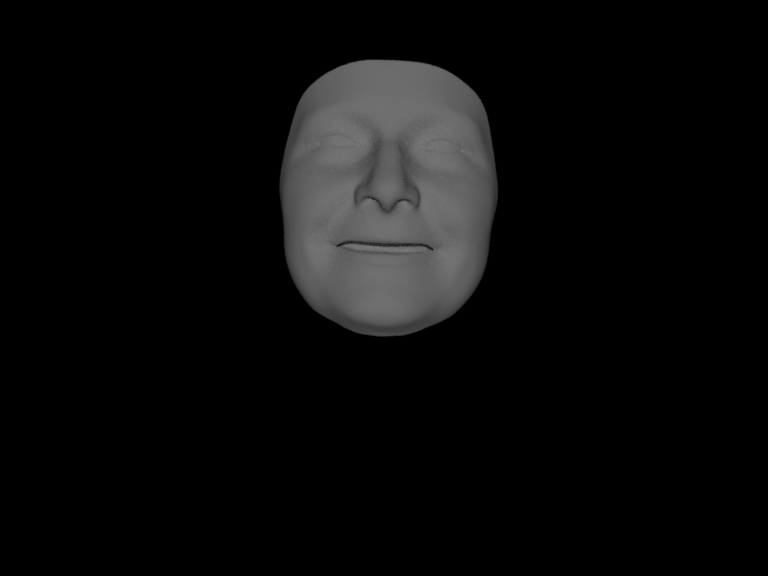} 
    \includegraphics[width=0.19\linewidth,trim={130px 50px 130px 50px},clip]{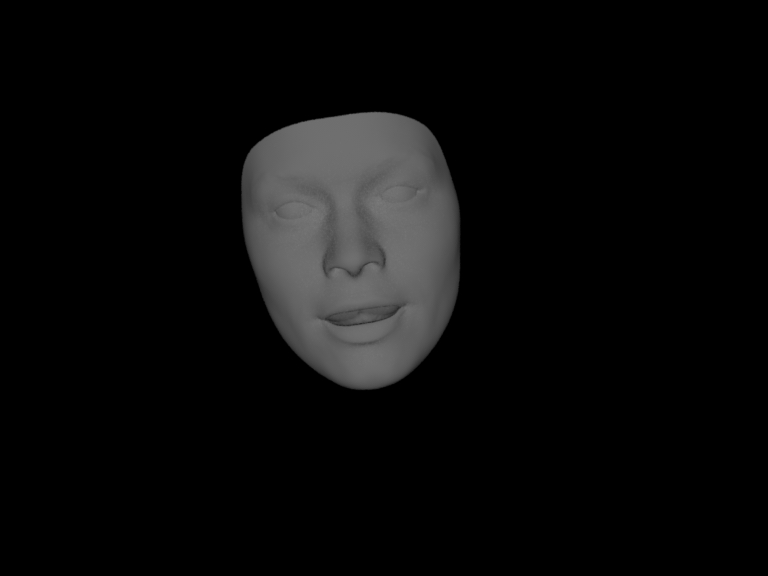} 
    \includegraphics[width=0.19\linewidth,trim={190 50px 70px 50px},clip]{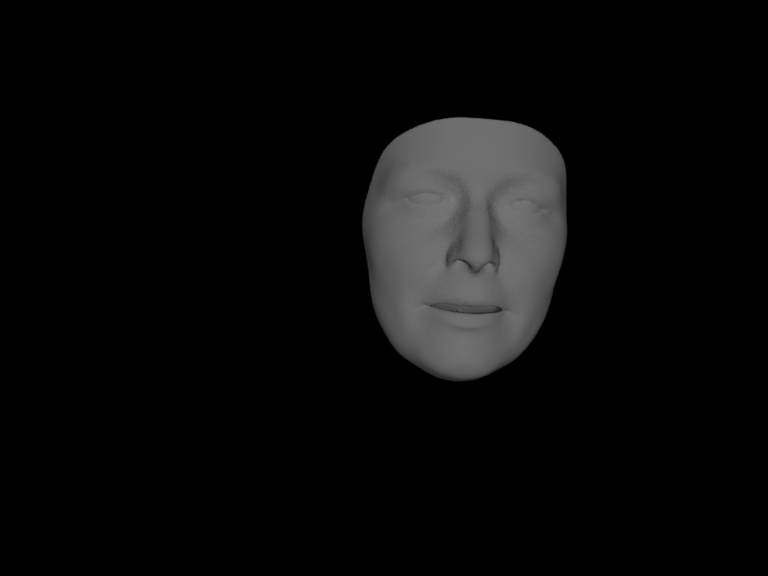} 
    \includegraphics[width=0.19\linewidth,trim={130px 50px 130px 50px},clip]{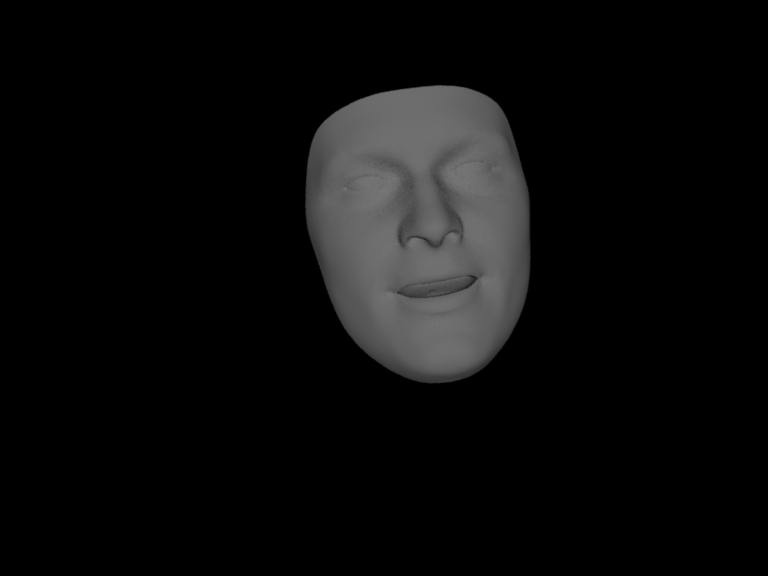}
    \end{subfigure}
    \end{minipage}
    \\
    & \begin{minipage}{\linewidth}
    \rotatebox{90}{Amp.-Depth}
    \end{minipage} & 
        \begin{minipage}{\linewidth}
    \begin{subfigure}{\linewidth}
    \includegraphics[width=0.19\linewidth,trim={130px 50px 130px 50px},clip]{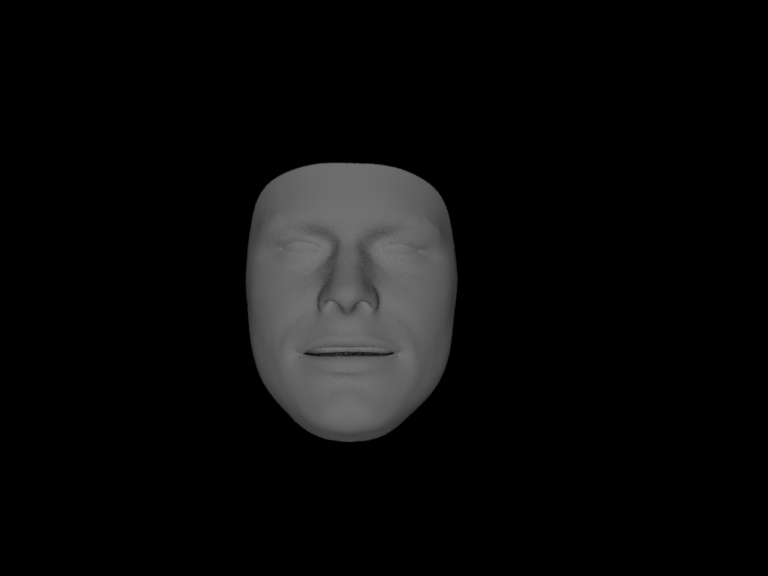} 
    \includegraphics[width=0.19\linewidth,trim={130px 100px 130px 0px},clip]{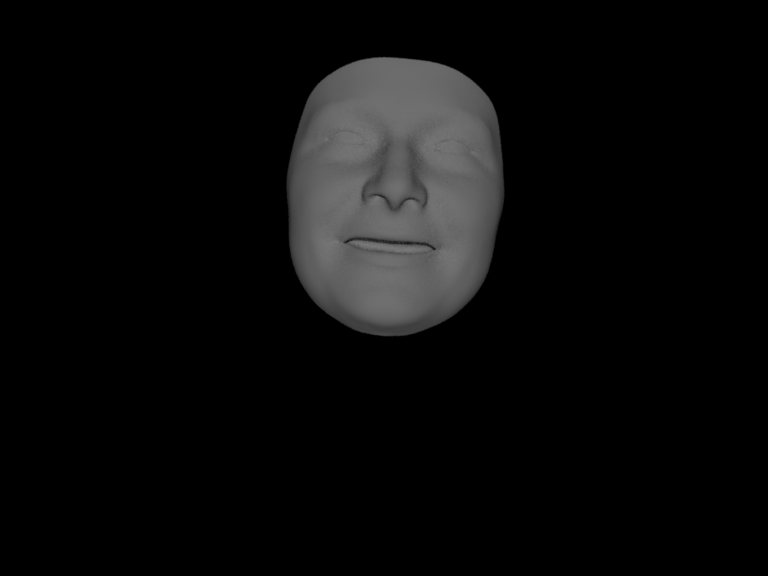} 
    \includegraphics[width=0.19\linewidth,trim={130px 50px 130px 50px},clip]{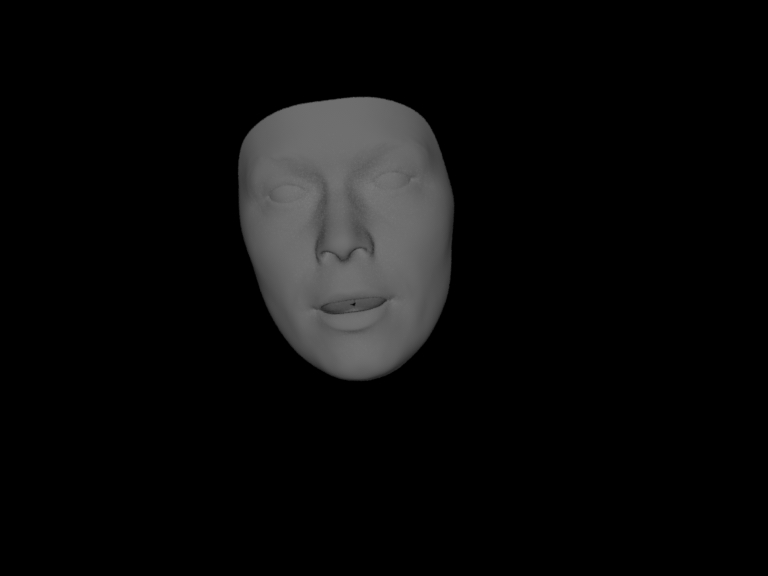} 
    \includegraphics[width=0.19\linewidth,trim={190 50px 70px 50px},clip]{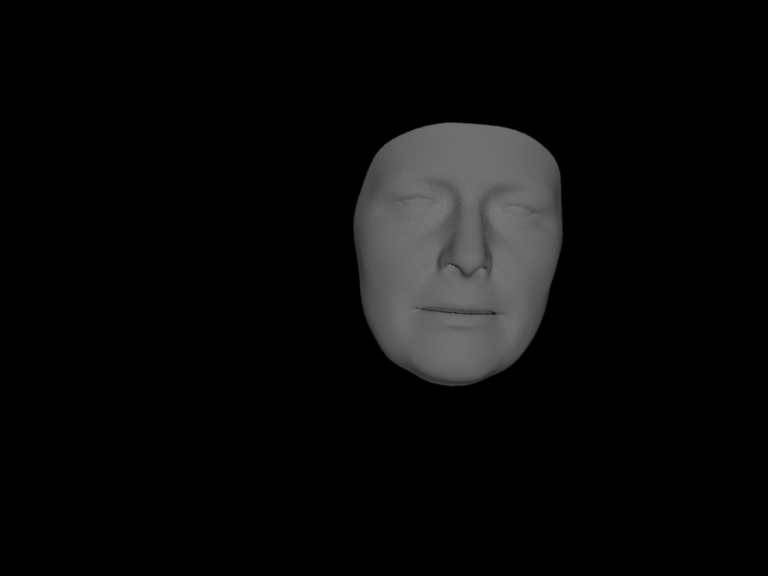} 
    \includegraphics[width=0.19\linewidth,trim={130px 50px 130px 50px},clip]{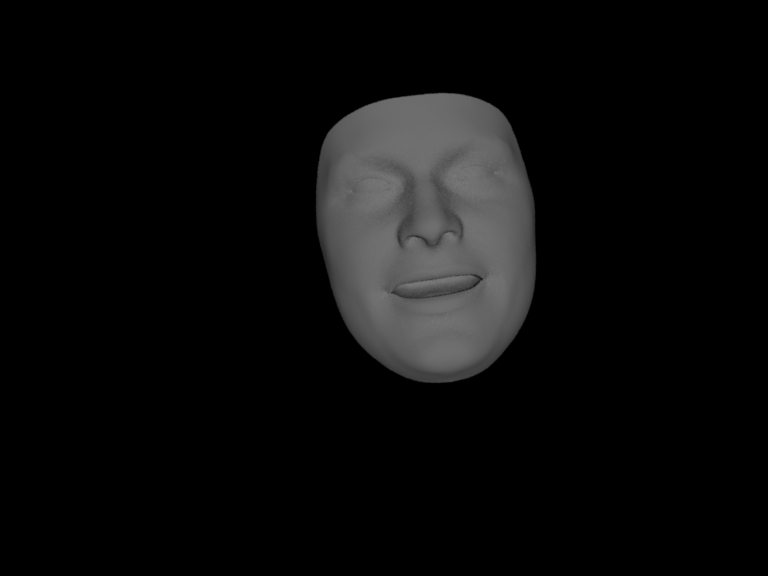} 
    \end{subfigure}
    \end{minipage}
    \\
  \end{tabular}
   \renewcommand{\@captype}{figure} 
  \caption{Mesh reconstructions of the results from the \textbf{encoder} and \textbf{autoencoder} models evaluated on \textbf{synthetic radar images}. Each column shows one face instance. }
  \label{fig:sup_mesh_synth}
\end{figure*}
\makeatother

\makeatletter
\begin{figure*}[h!]
  \centering
  \begin{tabular}{p{0.015\linewidth} p{0.015\linewidth} p{0.9\linewidth}}
    & 
    \begin{minipage}{\linewidth}
    \rotatebox{90}{Input}
    \end{minipage} & 
    \begin{minipage}{\linewidth}
    \begin{subfigure}{\linewidth}
        \includegraphics[width=0.19\linewidth,trim={130px 50px 130px 50px},clip]{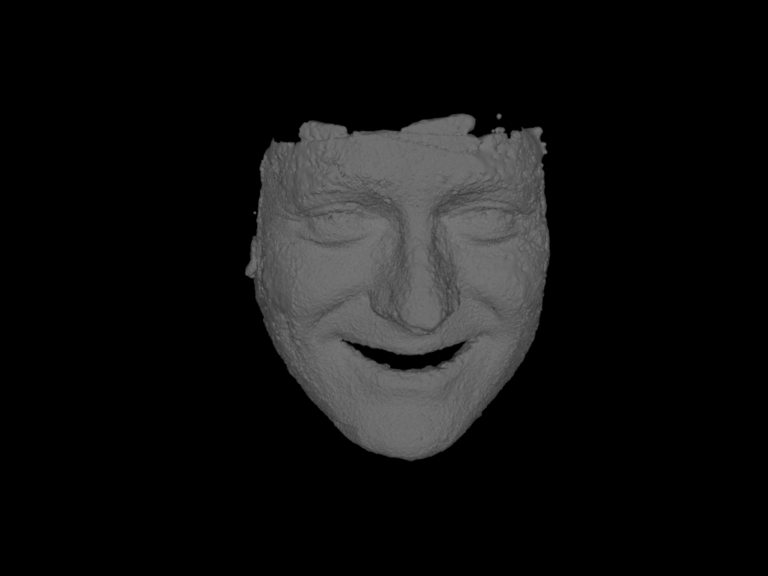} 
        \includegraphics[width=0.19\linewidth,trim={130px 50px 130px 50px},clip]{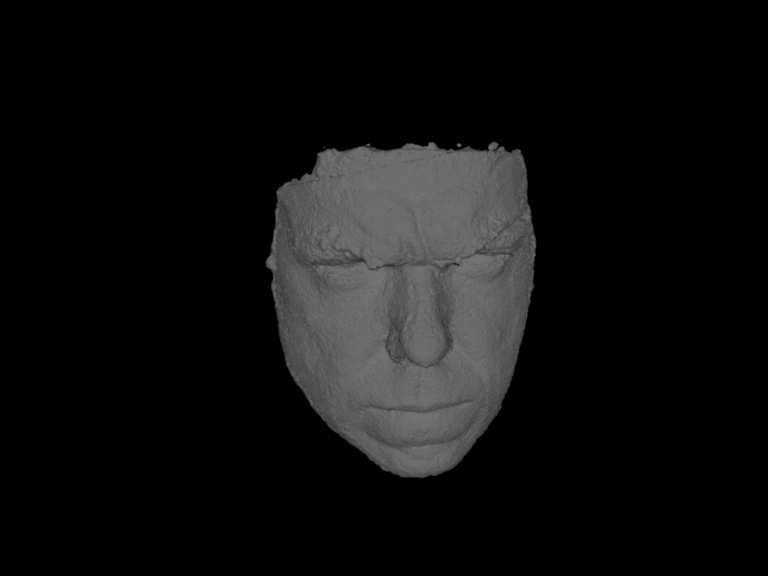} 
        \includegraphics[width=0.19\linewidth,trim={130px 50px 130px 50px},clip]{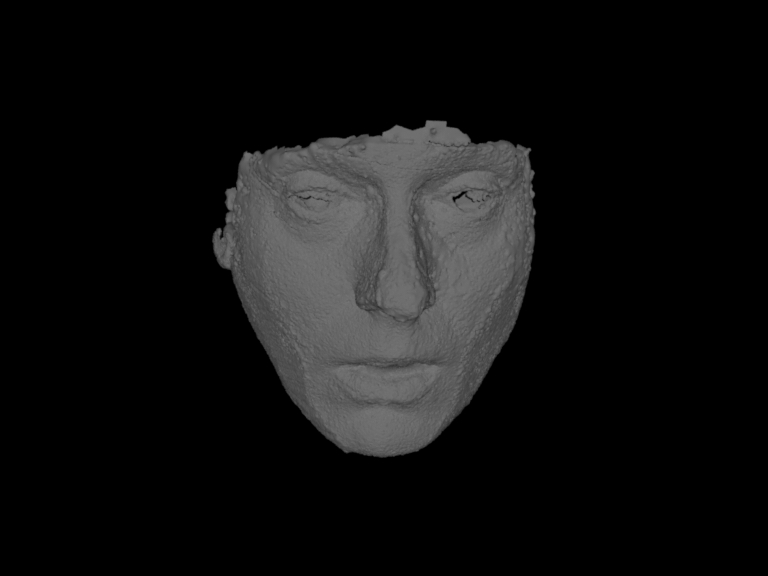} 
        \includegraphics[width=0.19\linewidth,trim={130px 50px 130px 50px},clip]{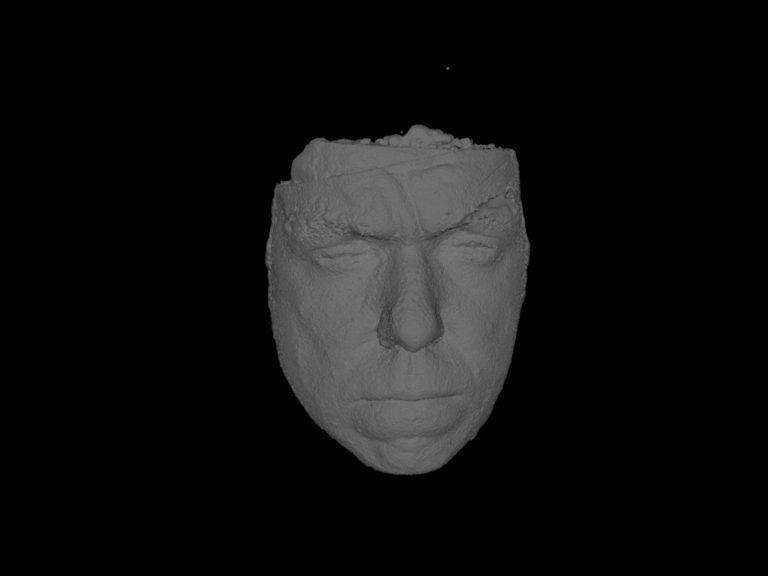} 
        \includegraphics[width=0.19\linewidth,trim={130px 50px 130px 50px},clip]{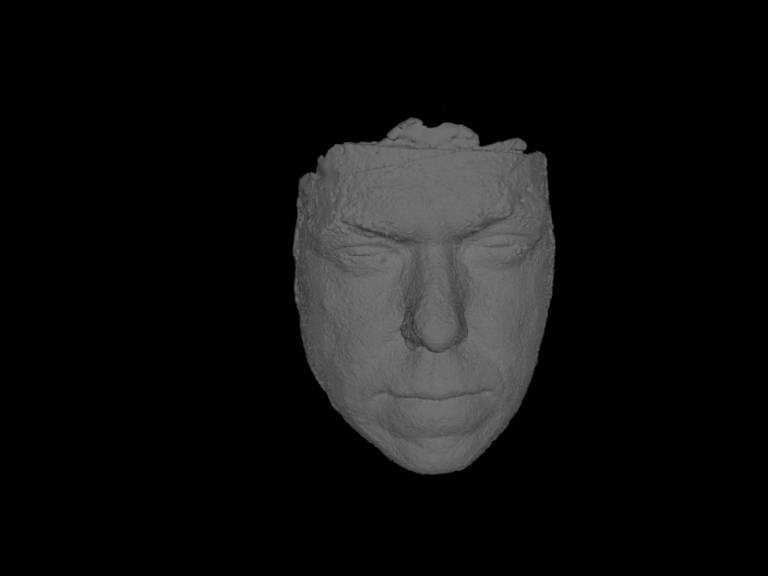} 
    \end{subfigure}
    \end{minipage} 
    \\
    \midrule

      & 
    \begin{minipage}{\linewidth}
    \rotatebox{90}{Amplitude}
    \end{minipage}  & 
    \begin{minipage}{\linewidth}
    \begin{subfigure}{\linewidth}
    \includegraphics[width=0.19\linewidth,trim={150px 110px 150px 30px},clip]{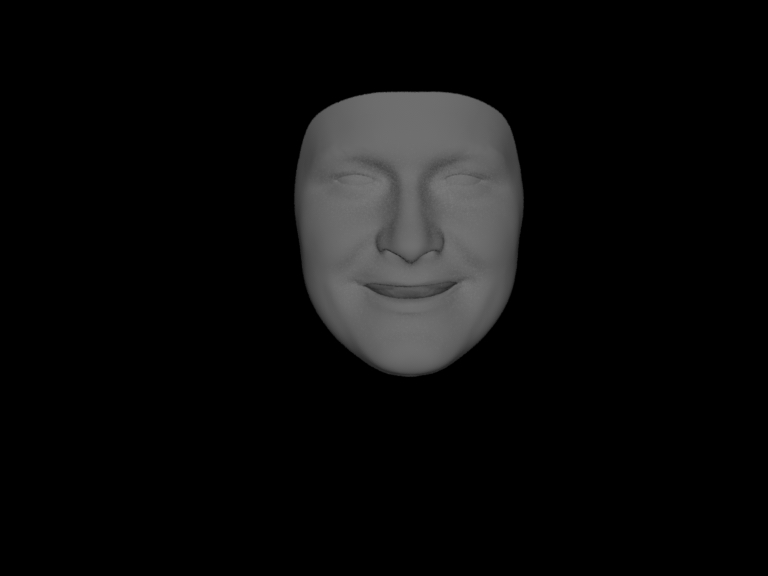} 
    \includegraphics[width=0.19\linewidth,trim={150px 110px 150px 30px},clip]{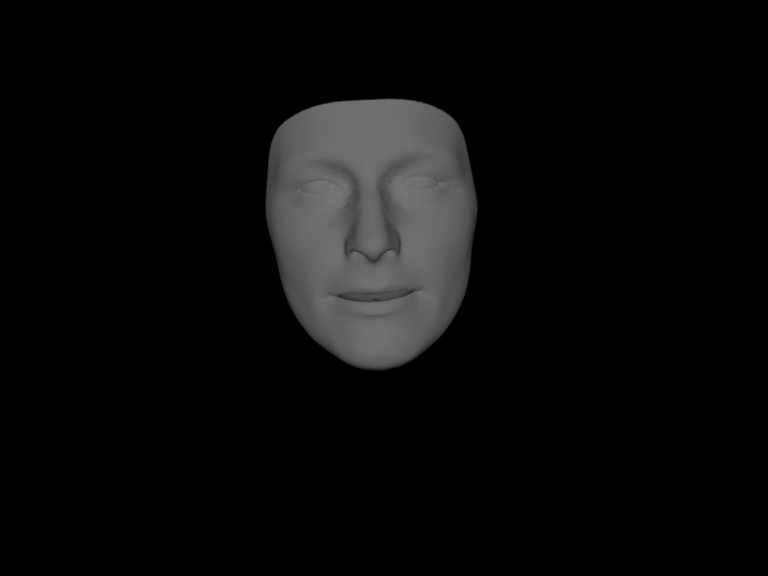} 
    \includegraphics[width=0.19\linewidth,trim={150px 110px 150px 30px},clip]{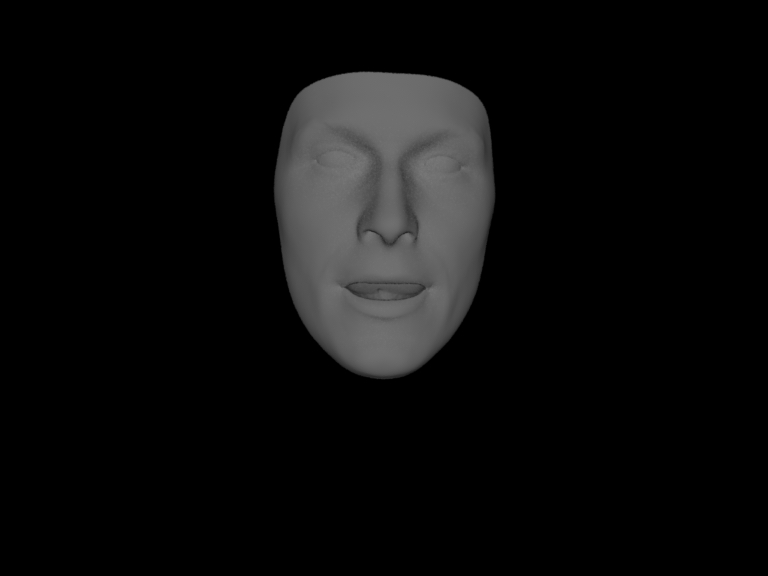} 
    \includegraphics[width=0.19\linewidth,trim={150px 110px 150px 30px},clip]{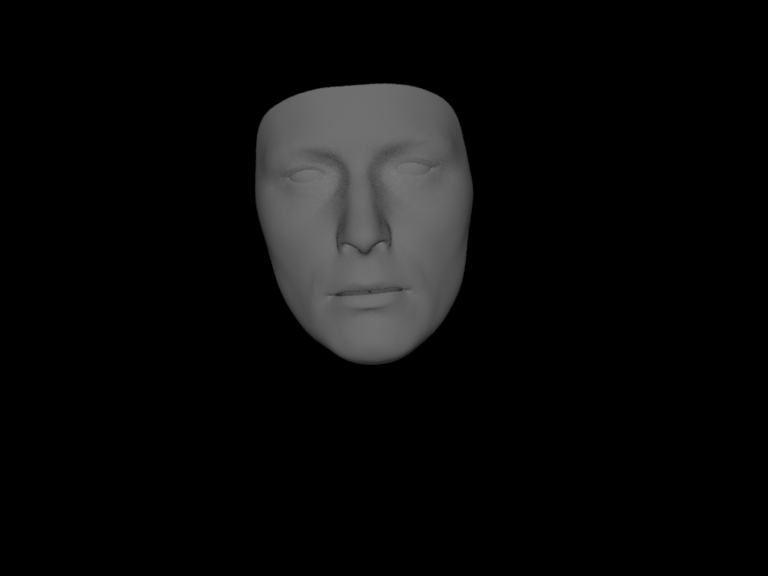} 
    \includegraphics[width=0.19\linewidth,trim={150px 110px 150px 30px},clip]{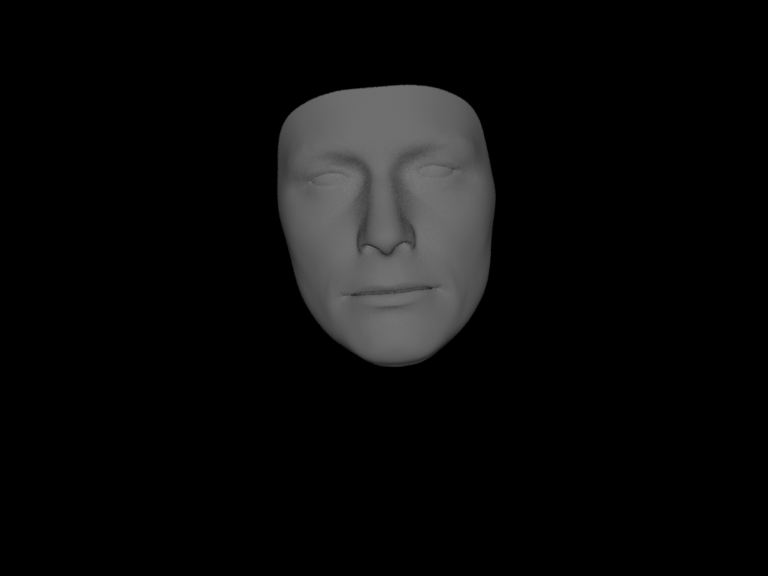}
    \end{subfigure}
    \end{minipage}
    \\
    \begin{minipage}{\linewidth}\rotatebox{90}{\textbf{Encoder}}\end{minipage}
    & \begin{minipage}{\linewidth}
    \rotatebox{90}{Depth}
    \end{minipage} & 
    \begin{minipage}{\linewidth}
    \begin{subfigure}{\linewidth}
    \includegraphics[width=0.19\linewidth,trim={150px 110px 150px 30px},clip]{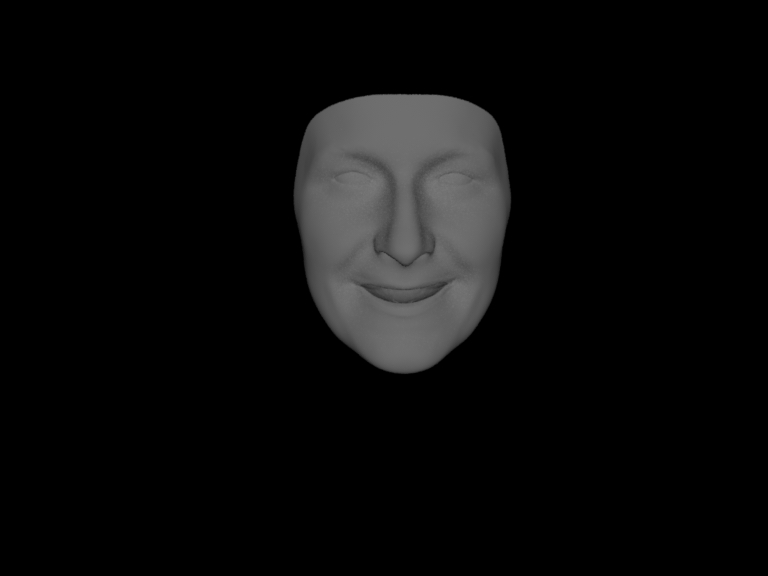} 
    \includegraphics[width=0.19\linewidth,trim={150px 110px 150px 30px},clip]{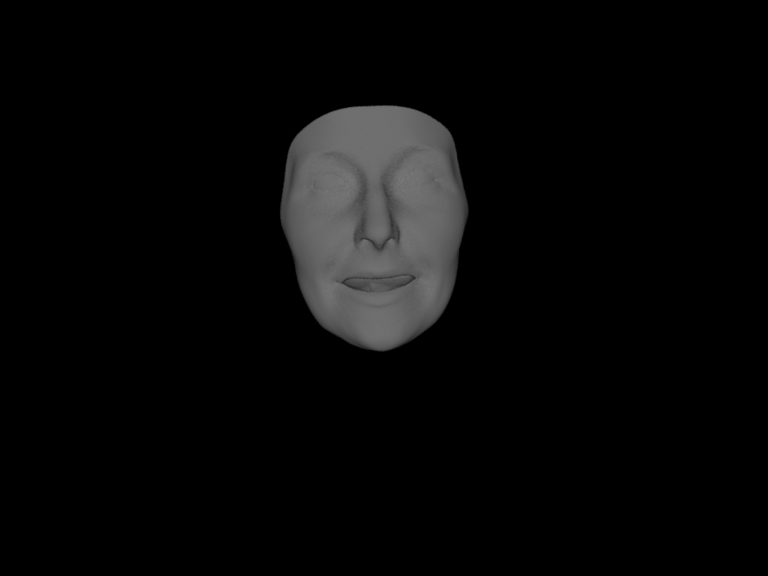} 
    \includegraphics[width=0.19\linewidth,trim={150px 110px 150px 30px},clip]{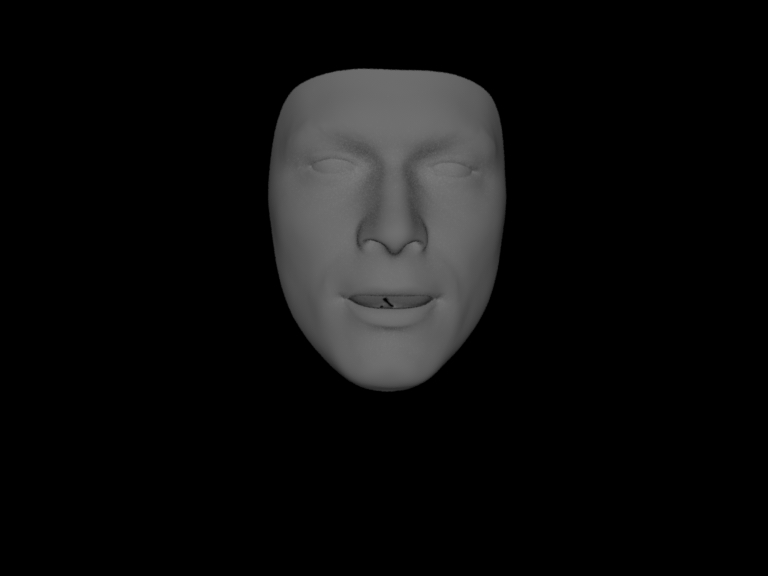} 
    \includegraphics[width=0.19\linewidth,trim={150px 110px 150px 30px},clip]{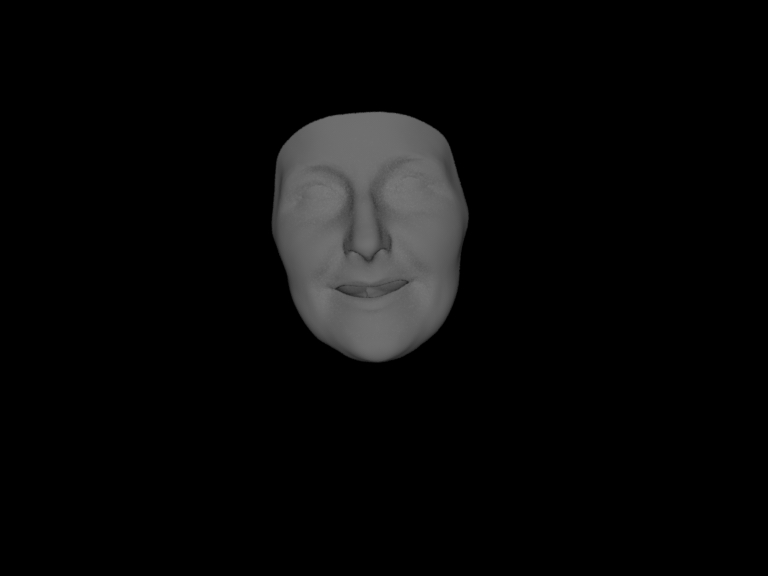} 
    \includegraphics[width=0.19\linewidth,trim={150px 110px 150px 30px},clip]{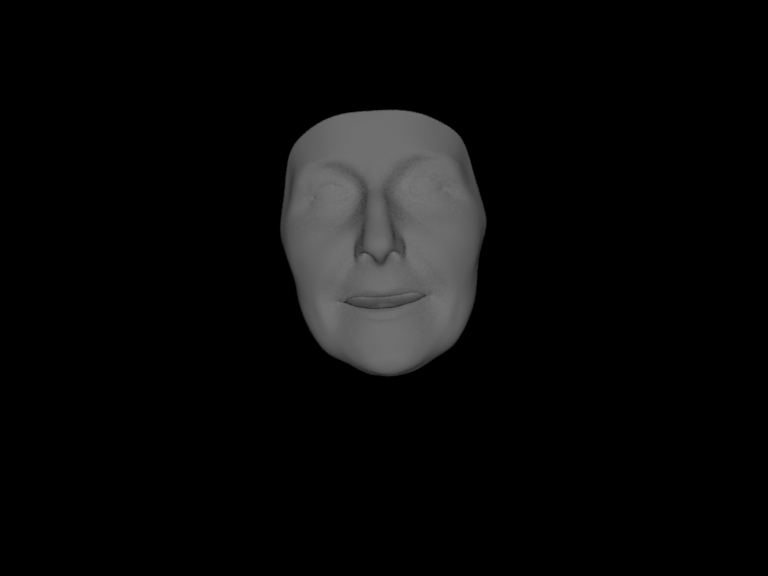}
    \end{subfigure}
    \end{minipage}
    \\
    & \begin{minipage}{\linewidth}
    \rotatebox{90}{Amp.-Depth}
    \end{minipage} & 
        \begin{minipage}{\linewidth}
    \begin{subfigure}{\linewidth}
    \includegraphics[width=0.19\linewidth,trim={150px 110px 150px 30px},clip]{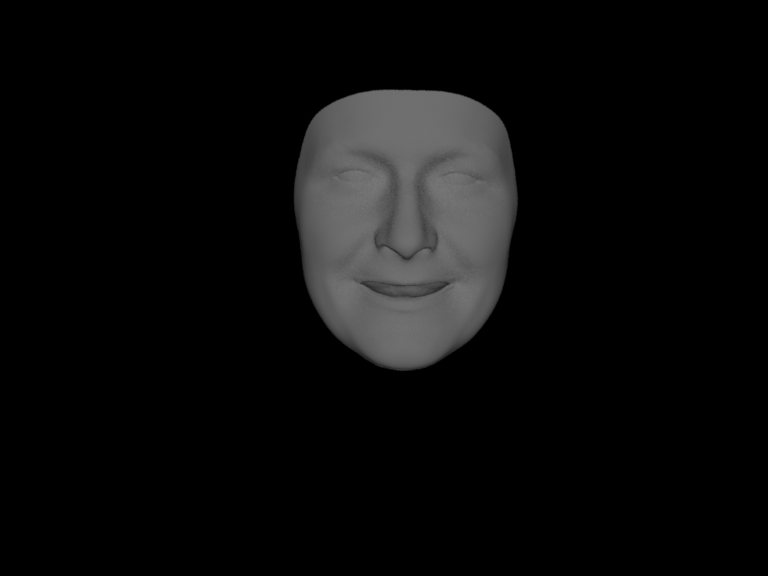}  
    \includegraphics[width=0.19\linewidth,trim={150px 110px 150px 30px},clip]{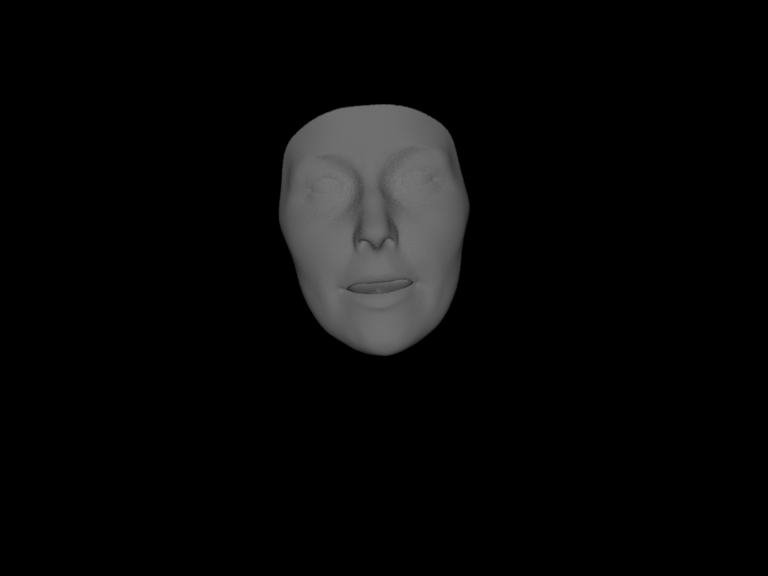}  
    \includegraphics[width=0.19\linewidth,trim={150px 110px 150px 30px},clip]{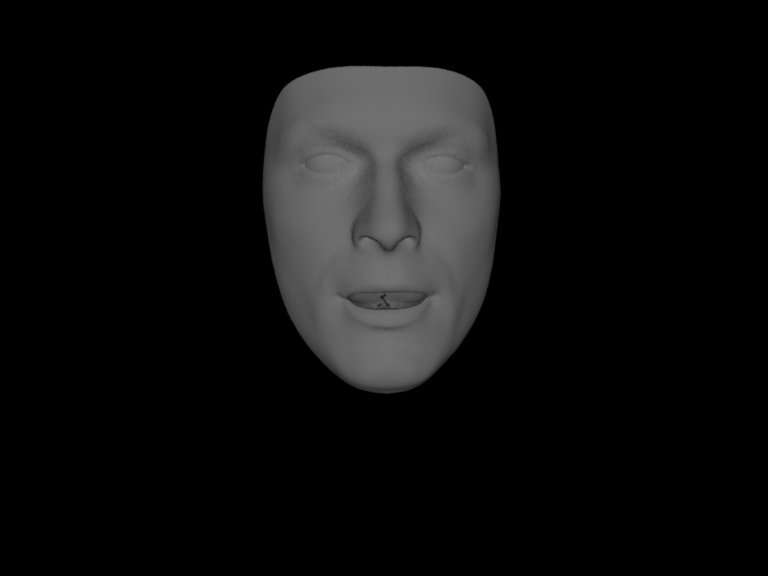}  
    \includegraphics[width=0.19\linewidth,trim={150px 110px 150px 30px},clip]{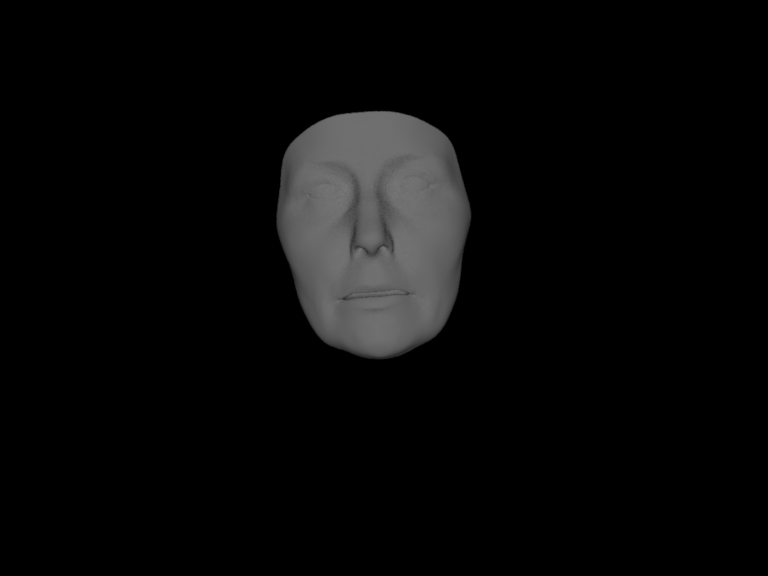}  
    \includegraphics[width=0.19\linewidth,trim={150px 110px 150px 30px},clip]{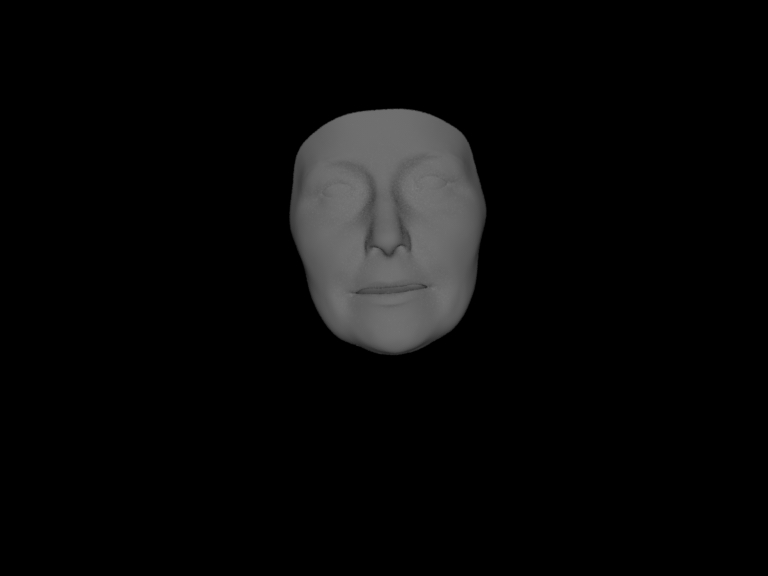}
    \end{subfigure}
    \end{minipage}
    \\
    \midrule
    & 
    \begin{minipage}{\linewidth}
    \rotatebox{90}{Amplitude}
    \end{minipage} & 
        \begin{minipage}{\linewidth}
    \begin{subfigure}{\linewidth}
    \includegraphics[width=0.19\linewidth,trim={150px 110px 150px 30px},clip]{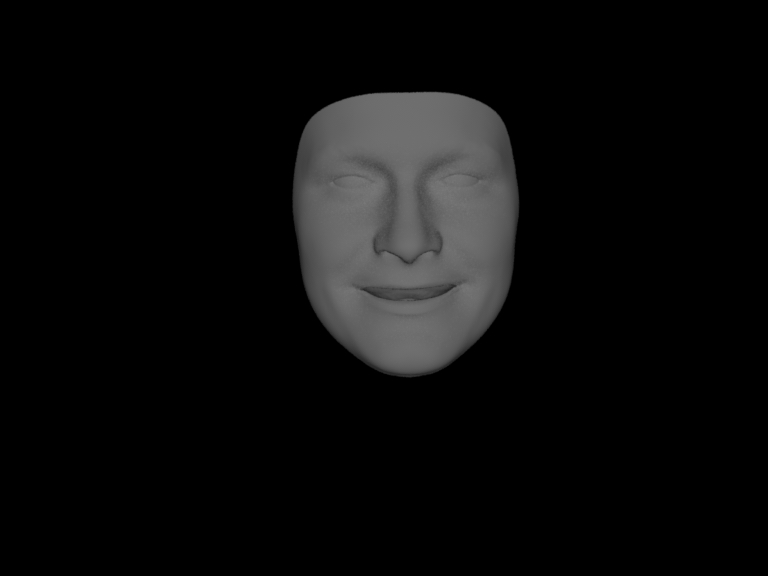} 
    \includegraphics[width=0.19\linewidth,trim={150px 110px 150px 30px},clip]{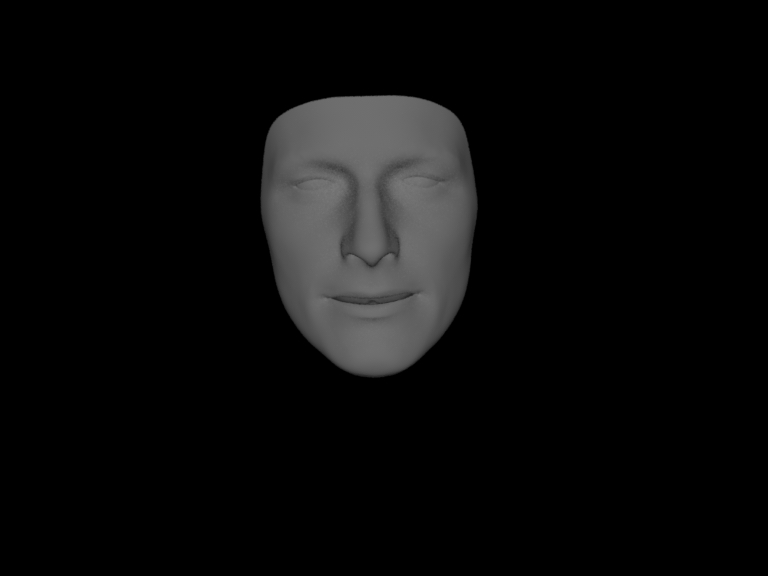} 
    \includegraphics[width=0.19\linewidth,trim={150px 110px 150px 30px},clip]{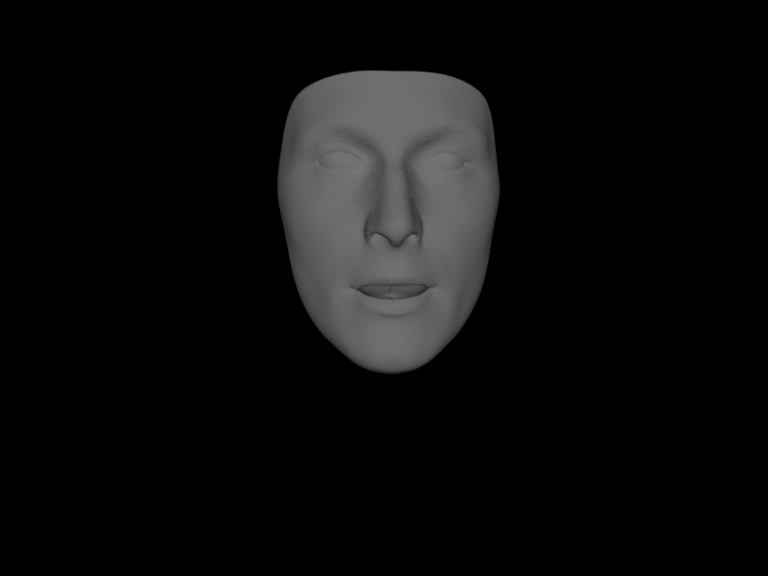}
    \includegraphics[width=0.19\linewidth,trim={150px 110px 150px 30px},clip]{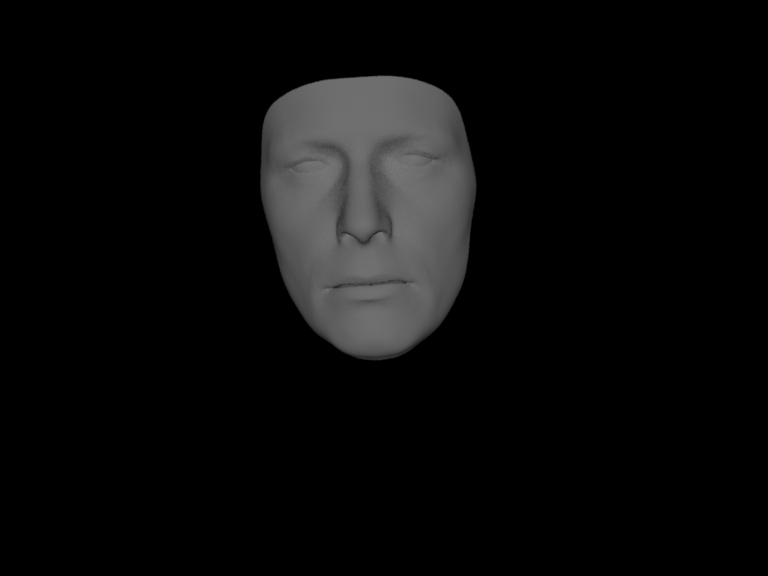} 
    \includegraphics[width=0.19\linewidth,trim={150px 110px 150px 30px},clip]{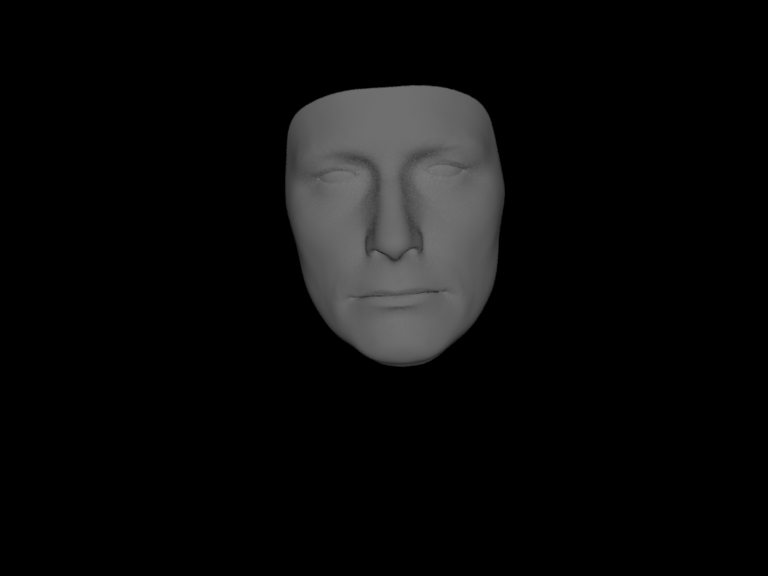}
    \end{subfigure}
    \end{minipage}
    \\
    \begin{minipage}{\linewidth}\rotatebox{90}{\textbf{Autoencoder}}\end{minipage} & \begin{minipage}{\linewidth}
    \rotatebox{90}{Depth}
    \end{minipage} & 
        \begin{minipage}{\linewidth}
    \begin{subfigure}{\linewidth}
    \includegraphics[width=0.19\linewidth,trim={150px 110px 150px 30px},clip]{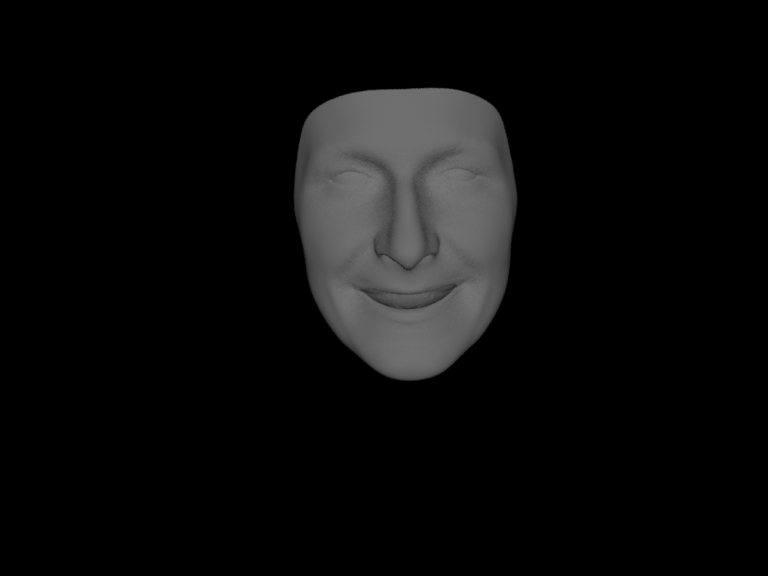} 
    \includegraphics[width=0.19\linewidth,trim={150px 110px 150px 30px},clip]{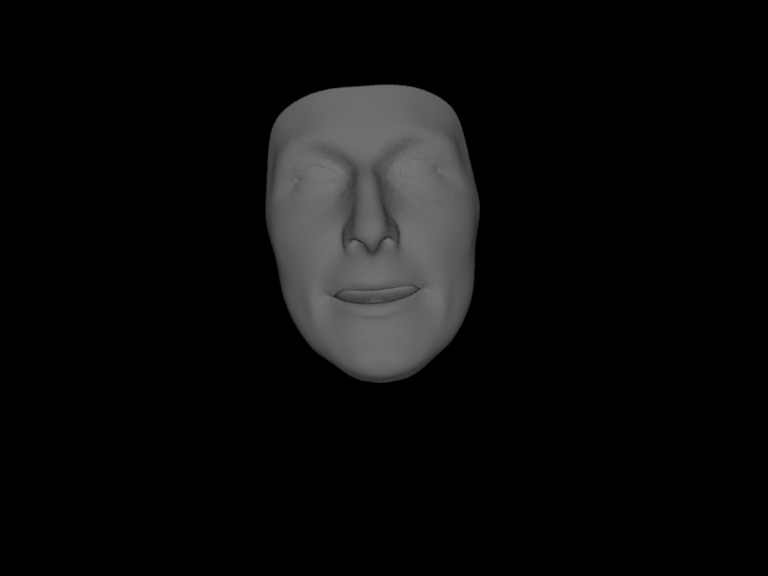} 
    \includegraphics[width=0.19\linewidth,trim={150px 110px 150px 30px},clip]{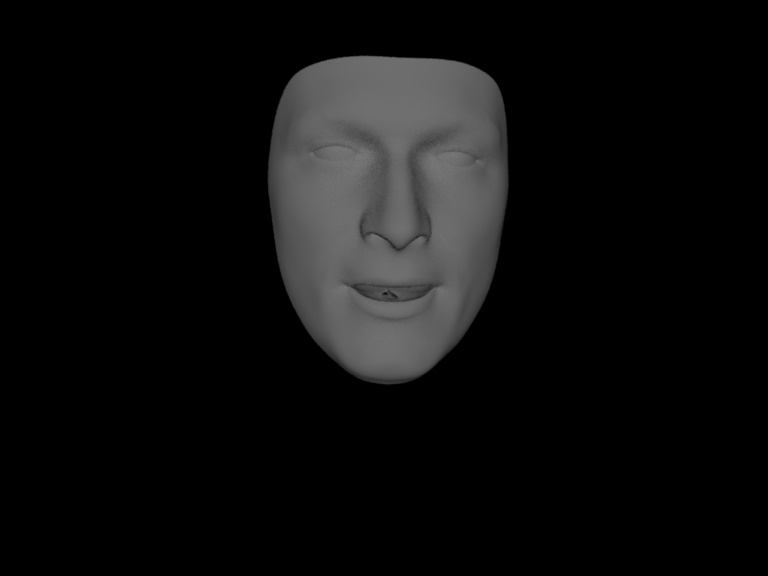} 
    \includegraphics[width=0.19\linewidth,trim={150px 110px 150px 30px},clip]{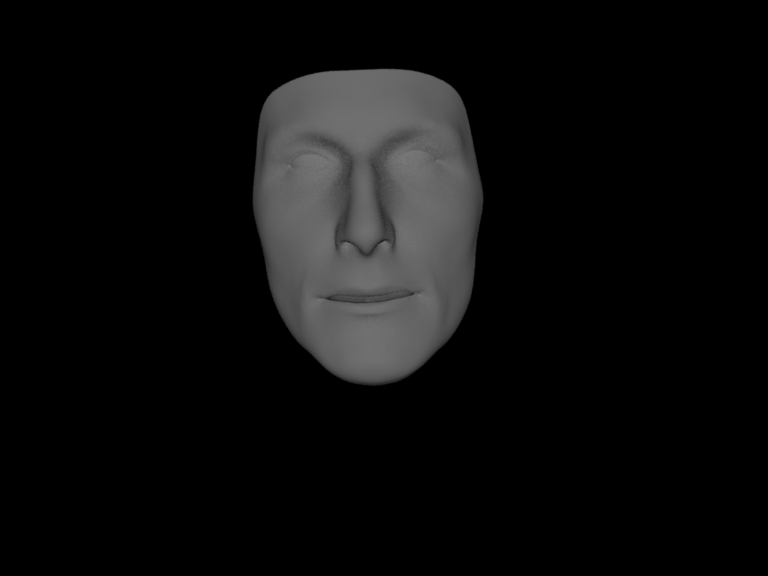} 
    \includegraphics[width=0.19\linewidth,trim={150px 110px 150px 30px},clip]{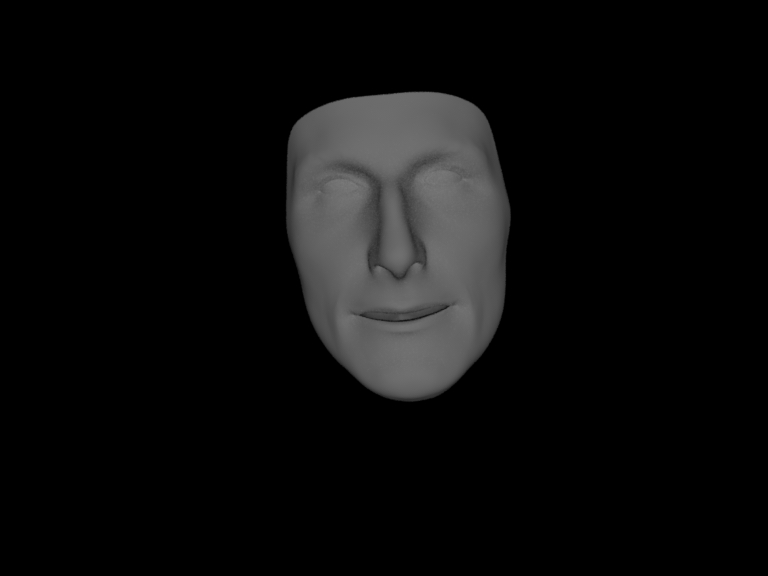}
    \end{subfigure}
    \end{minipage}
    \\
    & \begin{minipage}{\linewidth}
    \rotatebox{90}{Amp.-Depth}
    \end{minipage} & 
        \begin{minipage}{\linewidth}
    \begin{subfigure}{\linewidth}
    \includegraphics[width=0.19\linewidth,trim={150px 110px 150px 30px},clip]{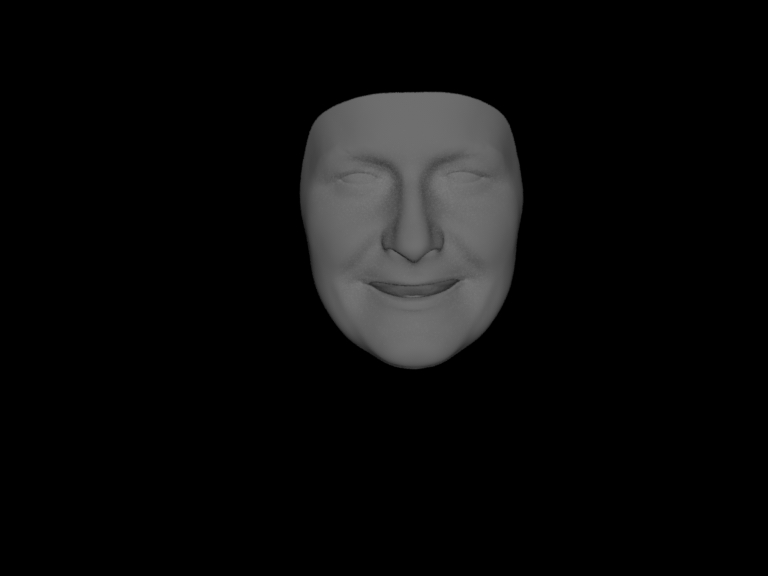} 
    \includegraphics[width=0.19\linewidth,trim={150px 110px 150px 30px},clip]{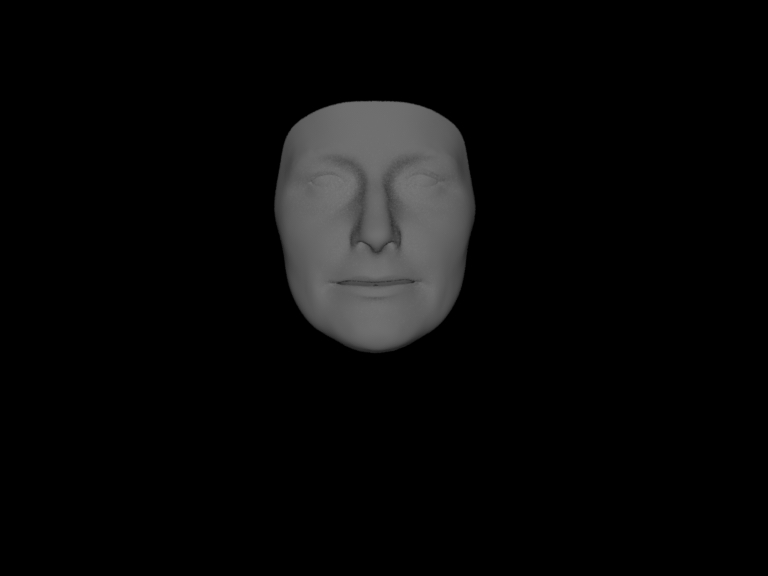} 
    \includegraphics[width=0.19\linewidth,trim={150px 110px 150px 30px},clip]{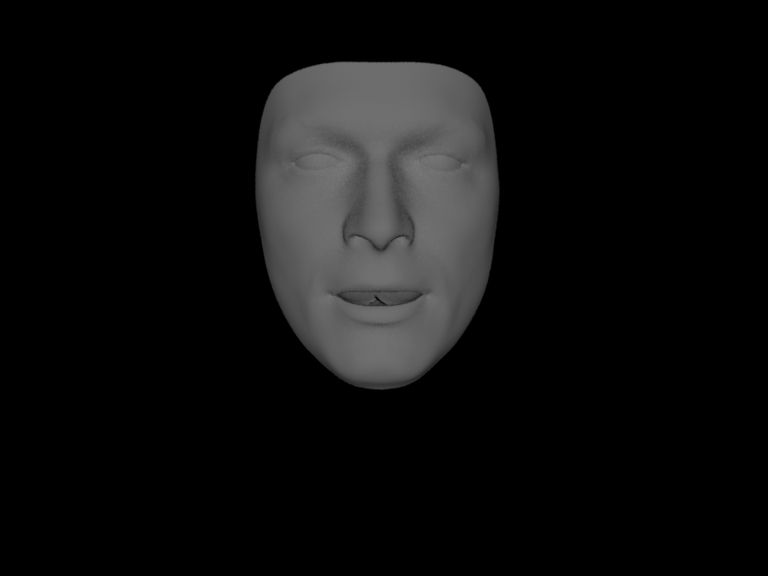} 
    \includegraphics[width=0.19\linewidth,trim={150px 110px 150px 30px},clip]{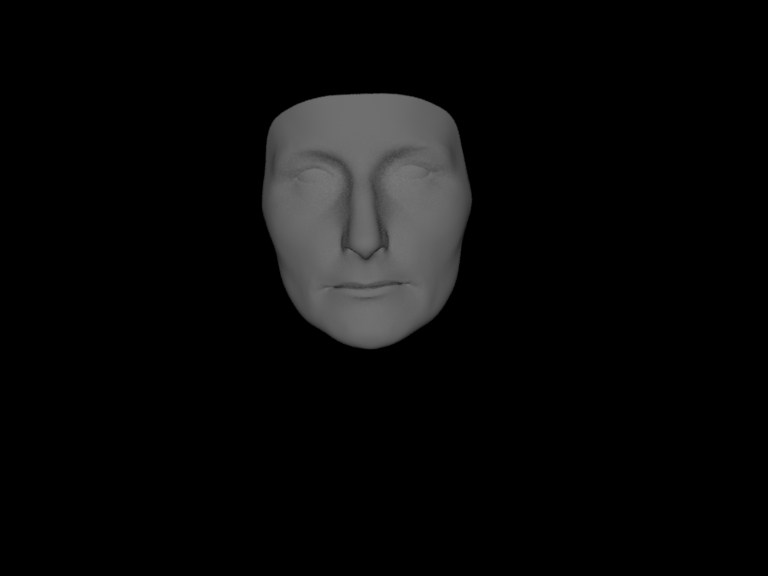} 
    \includegraphics[width=0.19\linewidth,trim={150px 110px 150px 30px},clip]{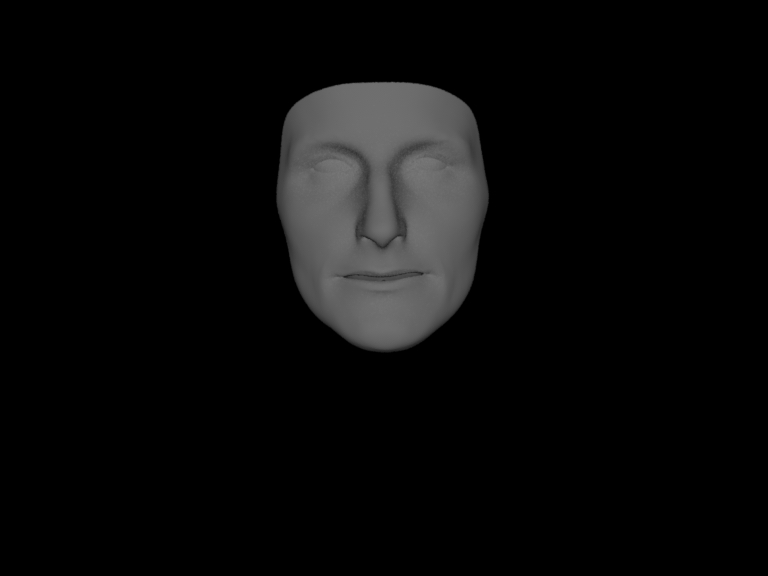} 
    \end{subfigure}
    \end{minipage}
    \\
  \end{tabular}
   \renewcommand{\@captype}{figure} 
  \caption{Mesh reconstructions of the results from the \textbf{encoder} and \textbf{autoencoder} models evaluated on \textbf{real radar images}. Each column shows one of the captured real persons with a different facial expression. }
  \label{fig:sup_mesh_real}
\end{figure*}
\makeatother

\end{document}